%% file: main_arxiv.tex
\documentclass[10pt,twocolumn,letterpaper]{article}

\usepackage{iccv}
\usepackage{times}
\usepackage{epsfig}
\usepackage{graphicx}
\usepackage{amsmath}
\usepackage{amssymb}

\usepackage{pdfpages}
\usepackage{xcolor}
\usepackage{makecell}
\usepackage{pifont}
\usepackage{capt-of}
\usepackage{subcaption}
\usepackage{array,multirow}
\usepackage{booktabs}
\usepackage{pifont}
\usepackage[export]{adjustbox}
\usepackage{multirow}

\usepackage{cite}

\newcolumntype{L}[1]{>{\raggedright\let\newline\\\arraybackslash\hspace{0pt}}m{#1}}
\newcolumntype{C}[1]{>{\centering\let\newline\\\arraybackslash\hspace{0pt}}m{#1}}
\newcolumntype{R}[1]{>{\raggedleft\let\newline\\\arraybackslash\hspace{0pt}}m{#1}}
\newcommand{\cmark}{\ding{51}}%
\newcommand{\red}[1]{\textcolor{red}{#1}}

\def\eg{\emph{e.g.}}
\def\ie{\emph{i.e.}}

\def\etal{\emph{et al.}}

\input{math_commands.tex}

\usepackage[pagebackref=true,breaklinks=true,letterpaper=true,colorlinks,bookmarks=false]{hyperref}

\iccvfinalcopy 

\def\iccvPaperID{****} 

\ificcvfinal\fi

\begin{document}

\title{Learning Self-Similarity in Space and Time \\as Generalized Motion for Video Action Recognition}

\author{
Heeseung Kwon\thanks{Equal contribution.} \hspace{10mm}
Manjin Kim\footnotemark[1] \hspace{10mm}
Suha Kwak \hspace{10mm}
Minsu Cho \vspace{2mm} \\ 
Pohang University of Science and Technology (POSTECH), South Korea\\
{\tt\small \url{http://cvlab.postech.ac.kr/research/SELFY/}}
}

\maketitle
\ificcvfinal\thispagestyle{empty}\fi

\input{sections/0_abstract.tex}


\input{sections/1_introduction.tex}

\input{sections/2_related_work.tex}

\input{sections/3_approach.tex}

\input{sections/4_experiment.tex}

\input{sections/5_conclusion.tex}

\vspace{2mm}

\noindent \textbf{Acknowledgements.} This work is supported by Samsung Advanced Institute of Technology (SAIT), the NRF grant (NRF-2021R1A2C3012728), and the IITP grant (No.2019-0-01906, AI Graduate School Program - POSTECH) funded by Ministry of Science and ICT, Korea.

{\small
\bibliographystyle{ieee_fullname}
\bibliography{cvlab_cho}
}

\input{sections/6_supp.tex}

\end{document}

%% file: math_commands.tex
\usepackage{amsmath,amsfonts,bm}

\def\eqref#1{equation~\ref{#1}}

\def\1{\bm{1}}

\DeclareMathAlphabet{\mathsfit}{\encodingdefault}{\sfdefault}{m}{sl}
\SetMathAlphabet{\mathsfit}{bold}{\encodingdefault}{\sfdefault}{bx}{n}

\DeclareMathOperator*{\argmax}{arg\,max}

%% file: sections/0_abstract.tex
\begin{abstract}
Spatio-temporal convolution often fails to learn motion dynamics in videos and thus an effective motion representation is required for video understanding in the wild. In this paper, we propose a rich and robust motion representation based on {\em spatio-temporal self-similarity} (STSS). Given a sequence of frames, STSS represents each local region as similarities to its neighbors in space and time. By converting appearance features into relational values, it enables the learner to better recognize structural patterns in space and time. We leverage the whole volume of STSS and let our model learn to extract an effective motion representation from it. 
The proposed neural block, dubbed {\em SELFY}, can be easily inserted into neural architectures and trained end-to-end without additional supervision. With a sufficient volume of the neighborhood in space and time, it effectively captures long-term interaction and fast motion in the video, leading to robust action recognition.
Our experimental analysis demonstrates its superiority over previous methods for motion modeling as well as its complementarity to spatio-temporal features from direct convolution. On the standard action recognition benchmarks, Something-Something-V1 \& V2, Diving-48, and FineGym, the proposed method achieves the state-of-the-art results.  
\end{abstract}

%% file: sections/1_introduction.tex

\begin{figure}[h]
    \captionsetup[subfigure]{position=b}
    \centering
    \includegraphics[width=\columnwidth]{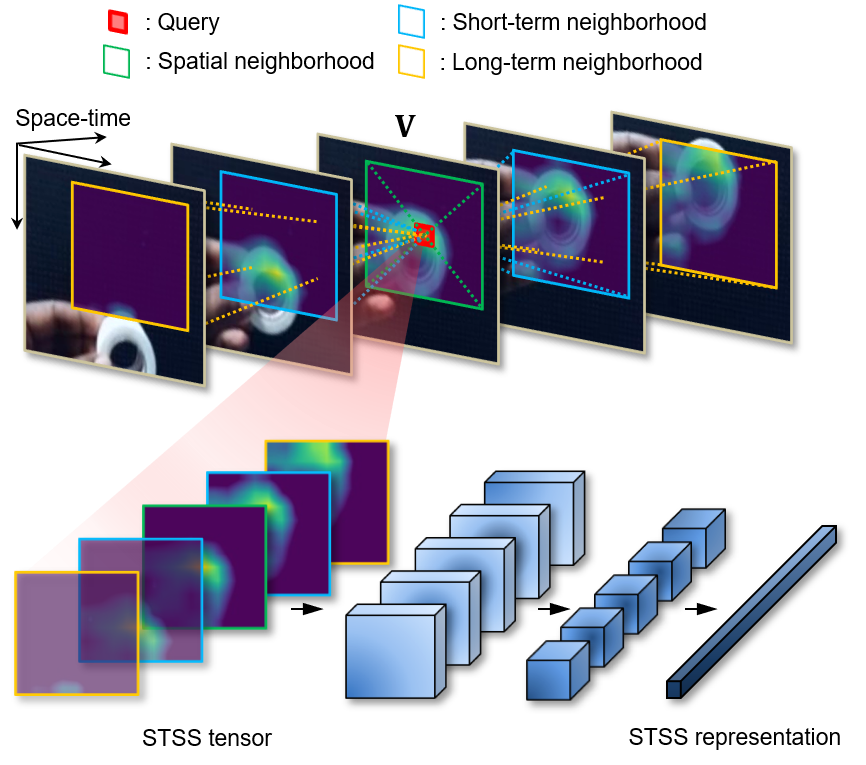}

\caption{\textbf{Spatio-temporal self-similarity (STSS) representation learning.} STSS describes each position (query) by its similarities (STSS tensor) with its neighbors in space and time (neighborhood). It allows to take a generalized, far-sighted view on motion, \ie, both short-term and long-term, both forward and backward, as well as spatial self-motion. Our method learns to extract a rich motion representation from STSS without additional supervision.  
} \label{fig:1}\vspace{-0.2cm}
\end{figure}

\section{Introduction}

Learning spatio-temporal dynamics is the key to video understanding. While extending standard convolution in space and time has been actively investigated for the purpose in recent years~\cite{tran2015learning,carreira2017quo,tran2018closer}, the empirical results so far indicate that spatio-temporal convolution alone is not sufficient for grasping the whole picture; 
it often learns irrelevant context bias rather than motion information~\cite{moon2021integralaction, materzynska2020something} and thus the additional use of optical flow turns out to boost the performance in most cases~\cite{carreira2017quo,lin2019tsm}. Motivated by this, recent action recognition methods learn to extract explicit motion, \ie, flow or correspondence, between feature maps of adjacent frames to improve the performance~\cite{li2020tea,kwon2020motionsqueeze}.  
But, is it essential to extract such an explicit form of flows or correspondences? How can we learn a richer and more robust form of motion information for videos in the wild?

In this paper, we propose to learn {\it spatio-temporal self-similarity} (STSS) representation for video understanding. 
Self-similarity is a relational descriptor for an image that effectively captures intra-structures by representing each local region as similarities to its spatial neighbors~\cite{shechtman2007matching}. As illustrated in Fig.~\ref{fig:1}, given a sequence of frames, \ie, a video, it extends along time and thus represents each local region as similarities to its neighbors in space and time. By converting appearance features into relational values, STSS enables a learner to better recognize structural patterns in space and time. 
For neighbors at the same frame it computes a spatial self-similarity map, while for neighbors at a different frame it extracts a motion likelihood map. 
Note that if we fix our attention to the similarity map to the very next frame within STSS and attempt to extract a single displacement vector to the most likely position at the frame, the problem reduces to optical flow, which is a limited type of motion information. In contrast, we leverage the whole volume of STSS and let our model learn to extract a generalized motion representation from it in an end-to-end manner. With a sufficient volume of the neighborhood in space and time, it effectively captures long-term interaction and fast motion in the video, leading to robust action recognition. 

We introduce a neural block for STSS representation, dubbed {\em SELFY}, that can be easily inserted into neural architectures and learned end-to-end without additional supervision. Our experimental analysis demonstrates its superiority over previous methods for motion modeling as well as its complementarity to spatio-temporal features from direct convolutions. On the standard benchmarks for action recognition, Something-Something V1\&V2~\cite{goyal2017something}, Diving-48~\cite{li2018resound}, and FineGym~\cite{shao2020finegym}, the proposed method achieves the state-of-the-art results.

%% file: sections/2_related_work.tex

\section{Related Work}

\noindent \textbf{Video action recognition}.
Video action recognition aims to categorize videos into pre-defined action classes and one of the main issues in action recognition is to capture temporal dynamics in videos.
For modern neural networks, previous methods attempt to learn temporal dynamics in different ways: two-stream networks with external optical flows~\cite{simonyan2014two,wang2016temporal}, recurrent networks~\cite{donahue2015long}, temporal pooling methods~\cite{girdhar2017attentional,kwon2018first}, and 3D CNNs~\cite{tran2015learning,carreira2017quo}.
Recent methods have introduced the advanced 3D CNNs~\cite{tran2018closer,tran2019video,feichtenhofer2020x3d,lin2019tsm,fan2020rubiksnet} and showed the effectiveness of capturing spatio-temporal features, so that 3D CNNs now become a \textit{de facto} approach to learn temporal dynamics in the video.
However, spatio-temporal convolution is vulnerable unless relevant features are well aligned across frames within the fixed-sized kernel.
To address this issue, a few methods adaptively translate the kernel offsets with deformable convolutions~\cite{zhao2018trajectory,li2020spatio}, while several methods~\cite{feichtenhofer2019slowfast,li2020smallbignet} modulate the other hyper-parameters, \eg, higher frame rate or larger spatial receptive fields.
Unlike these methods, we address the problem of the spatio-temporal convolution by a sufficient volume of STSS, capturing far-sighted spatio-temporal relations.

\noindent \textbf{Learning motion features.}
Since using the external optical flow benefits 3D CNNs to improve the action recognition accuracy~\cite{carreira2017quo,zolfaghari2018eco,tran2018closer}, several methods propose to learn frame-by-frame motion features from RGB sequences inside neural architectures.
Some methods~\cite{fan2018end,piergiovanni2019representation} internalize TV-L1~\cite{zach2007duality} optical flows into the CNN.
Frame-wise feature differences~\cite{sun2018optical,lee2018motion,jiang2019stm,li2020tea} are also utilized as the motion features. 
Recent correlation-based methods~\cite{wang2020video,kwon2020motionsqueeze} adopt a correlation operator~\cite{dosovitskiy2015flownet,sun2018pwc,yang2019volumetric} to learn motion features between adjacent frames.
However, these methods compute frame-by-frame motion features between two adjacent frames and then rely on stacked spatio-temporal convolutions for capturing long-range motion dynamics.
In contrast, we propose to learn STSS features, as generalized motion features, that enable to capture both short-term and long-term interactions in the video.

\noindent \textbf{Self-similarity.}
Self-similarity describes a relational structure of individual image features by computing similarities between them~\cite{shechtman2007matching}.
Several methods~\cite{shechtman2007matching,junejo2010view,junejo2008cross,torabi2013local} use the self-similarity as a shallow relational descriptor, which is robust to photometric variations, in fields of template matching~\cite{shechtman2007matching}, capturing view-invariant geometric patterns~\cite{junejo2010view,junejo2008cross}, or finding semantic correspondences~\cite{torabi2013local,kim2017fcss,kang2021relational}.
In video understanding, there are a few approaches~\cite{wang2018non,liu2019learning} that use the self-similarity of a video as a form of STSS.
These methods, however, use STSS for a subsequent feature aggregation step rather than learn representation from it;
non-local operation~\cite{wang2018non} uses STSS as attention weights in aggregating features~\cite{vaswani2017attention, hu2019local,ramachandran2019stand,srinivas2021bottleneck} and CPNet~\cite{liu2019learning} uses STSS in selecting pairs of appearance features.
All these methods lose rich motion information of STSS during aggregation, being not suitable for capturing motion content of videos.
In contrast, we advocate using STSS directly for motion representation learning.
Our method leverages the full STSS as generalized motion information and learns an effective representation for action recognition within a video-processing architecture.
To the best of our knowledge, our work is the first in learning STSS representation using modern neural networks.

The contribution of our paper can be summarized as follows. 
First, we revisit the notion of self-similarity and propose to learn a generalized, far-sighted motion representation from STSS.
Second, we implement STSS representation learning as a neural block, dubbed {\em SELFY}, that can be integrated into existing neural architectures.
Third, we provide comprehensive evaluations on SELFY, achieving the state-of-the-art on benchmarks: Something-Something V1\&V2~\cite{goyal2017something}, Diving-48~\cite{li2018resound}, and FineGym~\cite{shao2020finegym}.

%% file: sections/3_approach.tex
\begin{figure*}[t]
    \centering
    \includegraphics[width=0.8\linewidth]{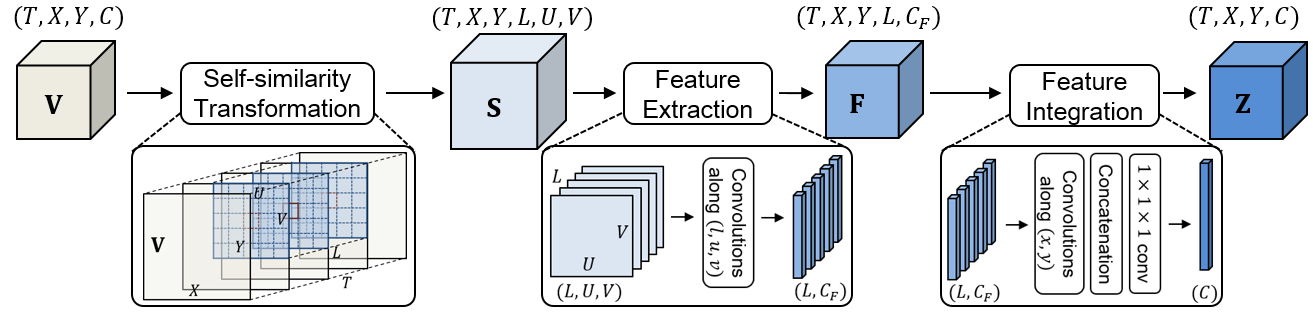}
\caption{\textbf{Overview of our self-similarity representation block (SELFY)}.
SELFY block takes as input a video feature tensor $\mathbf{V}$, transforms it to a STSS tensor $\mathbf{S}$, and extracts a feature tensor $\mathbf{F}$ from $\mathbf{S}$. It then produces the final STSS representation $\mathbf{Z}$ via the feature integration, which is the same size as the input $\mathbf{V}$. The resultant representation
$\mathbf{Z}$ is fused into the input feature $\mathbf{V}$ by element-wise addition, thus making SELFY act as a residual block. See text for details. 
} \label{fig:2}
\end{figure*}

\section{Our approach}
\label{headings}

In this section, we first revisit the notions of self-similarity and discuss its relation to motion. We then introduce our method for learning effective spatio-temporal self-similarity representation, which can be easily integrated into video-processing architectures and learned end-to-end.

\subsection{Self-Similarity Transformation}
\label{sec:spatio-temporal self-similarity}

Self-similarity is a relational descriptor that suppresses variations in appearance and reveals structural patterns~\cite{shechtman2007matching}. 

Given an image feature map $\mathbf{I} \in \mathbb{R}^{X\times Y\times C}$, self-similarity transformation of $\mathbf{I}$ results in a 4D tensor $\mathbf{S}\in \mathbb{R}^{X\times Y\times U\times V}$, whose elements are defined as 
\begin{align}
    \mathbf{S}_{x,y,u,v}=\mathrm{sim}(\mathbf{I}_{x,y}, \mathbf{I}_{x+u,y+v}),\nonumber    
\end{align}
where $\mathrm{sim}(\cdot,\cdot)$ is a similarity function, \eg, cosine similarity. 
Here, $(x,y)$ is a query coordinate while $(u,v)$ is a spatial offset from it.
To impose a locality, the offset is restricted to its neighborhood: $(u,v) \in [-d_\mathrm{U}, d_\mathrm{U}] \times [-d_\mathrm{V}, d_\mathrm{V}]$, so that $U=2d_\mathrm{U}+1$ and $V=2d_\mathrm{V}+1$, respectively.
By converting $C$-dimensional appearance feature $\mathbf{I}_{x,y}$ into $UV$-dimensional relational feature $\mathbf{S}_{x,y}$, it suppresses variations in appearance and reveals spatial structures in the image.
 Note that the self-similarity transformation closely relates to conventional cross-similarity (or correlation) across two different feature maps ($\mathbf{I}$, $\mathbf{I'} \in \mathbb{R}^{X\times Y\times C}$), which can be defined as
\begin{align}
    \mathbf{S}_{x,y,u,v}=\mathrm{sim}(\mathbf{I}_{x,y},\mathbf{I}'_{x+u,y+v}).\nonumber
\end{align} 
Given a moving object of two images, the cross-similarity transformation effectively captures motion information and thus is commonly used in optical flow and correspondence estimation~\cite{dosovitskiy2015flownet,sun2018pwc,yang2019volumetric}.

For a sequence of frames, \ie, a video, one can naturally extend the spatial self-similarity along the temporal axis.
Let  $\mathbf{V} \in \mathbb{R}^{T\times X\times Y\times C}$ denote a feature map of the video with $T$ frames. {\em Spatio-temporal self-similarity} (STSS) transformation of $\mathbf{V}$ results in a 6D tensor $\mathbf{S} \in \mathbb{R}^{T\times X\times Y\times L\times U\times V}$, whose elements are defined as 
\begin{align}
\label{eq:stss}
    \mathbf{S}_{t,x,y,l,u,v}=\mathrm{sim}(\mathbf{V}_{t,x,y}, \mathbf{V}_{t+l,x+u,y+v}), 
\end{align}
where $(t,x,y)$ is a query coordinate and $(l,u,v)$ is a spatio-temporal offset from the query.
In addition to the locality of spatial offsets above, the temporal offset $l$ is also restricted to its temporal neighborhood: $l \in [-d_\mathrm{L},d_\mathrm{L}]$, so that $L=2d_\mathrm{L}+1$.

What types of information does STSS describe? Interestingly, for each time $t$, the STSS tensor $\mathbf{S}$ can be decomposed along temporal offset $l$ into a single spatial self-similarity tensor (when $l = 0$) and $2d_\mathrm{L}$ spatial cross-similarity tensors (when $l \neq 0$); 
the partial tensors with a small offset (\eg, $l = -1$ or $+1$) collect motion information from adjacent frames and those with larger offsets capture it from further frames both forward and backward in time. 
Unlike previous approaches to learn motion~\cite{dosovitskiy2015flownet, wang2020video, kwon2020motionsqueeze}, which rely on cross-similarity between adjacent frames, STSS allows to take a generalized, far-sighted view on motion, \ie, both short-term and long-term, both forward and backward, as well as spatial self-motion. 

\begin{figure*}[t]
\captionsetup[subfigure]{position=b}
\centering
    \begin{subfigure}{.31\linewidth}
    \centering
    \includegraphics[width=\linewidth]{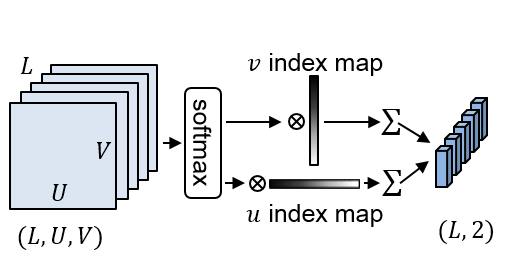}
    \caption{soft-argmax} 
    \label{fig:Soft_argmax}
    \end{subfigure}
    \hfill
    \begin{subfigure}{.36\linewidth}
    \centering
    \includegraphics[width=\linewidth]{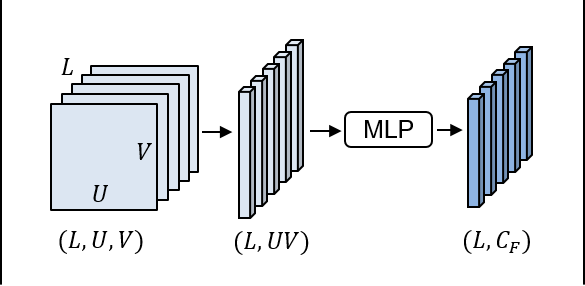}
    \caption{MLP} 
    \label{fig:mlp}
    \end{subfigure}
    \hfill
    \begin{subfigure}{.275\linewidth}
    \centering
    \includegraphics[width=\linewidth]{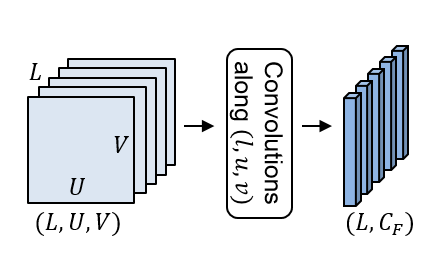}
    \caption{convolution} 
    \label{fig:conv}
    \end{subfigure}
\caption{\textbf{Feature extraction from STSS.} For a spatio-temporal position $(t,x,y)$, each method transforms $(L,U,V)$ volume of STSS tensor $\mathbf{S}$ into $(L,C_F)$. See text for details.} \label{fig:3}
\end{figure*}

\subsection{Spatio-temporal self-similarity representation learning} \label{sec:learning_stss}
By leveraging the rich information in STSS, we propose to learn a generalized motion representation for video understanding. To achieve this goal without additional supervision, we design a neural block, dubbed SELFY, which can be inserted into video-processing architectures and learned end-to-end. Figure~\ref{fig:2} illustrates the overall structure. It consists of three steps: {\em self-similarity transformation}, {\em feature extraction}, and {\em feature integration}.

Given the input video feature tensor $\mathbf{V}$, the self-similarity transformation step converts it to the STSS tensor $\mathbf{S}$ as in Eq.~\ref{eq:stss}. In the following, we describe feature extraction and integration steps. 

\subsubsection{Feature extraction} \label{sec:instantiation}
From the STSS tensor $\mathbf{S} \in \mathbb{R}^{T\times X\times Y\times L\times U\times V}$, we extract a $C_F$-dimensional feature for each spatio-temporal position $(t,x,y)$ and temporal offset $l$ so that the resultant tensor is $\mathbf{F} \in \mathbb{R}^{T\times X\times Y\times L\times C_F}$, which is equivariant to translation in space, time, and temporal offset. The dimension of $L$ is preserved to extract motion information across different temporal offsets in a consistent manner.
While there exist many design choices, we introduce three methods for feature extraction in this work.

\noindent \textbf{Soft-argmax.}
The first method is to compute explicit displacement fields using $\mathbf{S}$, which previous motion learning methods adopt using spatial cross-similarity~\cite{dosovitskiy2015flownet,sun2018pwc,yang2019volumetric}.
One may extract the displacement field by indexing the positions with the highest similarity value via $\argmax_{(u,v)}$, but it is not differentiable. 
We instead use {\em soft-argmax}~\cite{chapelle2010gradient}, which aggregates displacement vectors with softmax weighting (Fig.~\ref{fig:Soft_argmax}). The soft-argmax feature extraction can be formulated as  
\begin{align}
    \mathbf{F}_{t,x,y,l}=\sum_{u,v}
    \frac{\exp(\mathbf{S}_{t,x,y,l,u,v} / \tau )}{\sum_{u',v'}{\exp(\mathbf{S}_{t,x,y,l,u',v'} / \tau )}} [ u; v ],
\end{align}
which results in a feature tensor $\mathbf{F} \in \mathbb{R}^{T\times X\times Y\times L\times 2}$. The temperature factor $\tau$ adjusts the softmax distribution, and we set $\tau=0.01$ in our experiments.

\noindent \textbf{Multi-layer perceptron (MLP).} 
The second method is to learn an MLP that converts self-similarity values into a feature.
For this, we flatten the $(U,V)$ volume into $UV$-dimensional vectors, and apply an MLP to them (Fig.~\ref{fig:mlp}).
For the reshaped tensor $\mathbf{S} \in \mathbb{R}^{T\times X\times Y\times L\times UV}$,
a perceptron $f(\cdot)$ can be expressed as 
\begin{align}
    f(\mathbf{S}) = \mathrm{ReLU}(\mathbf{S}\times_{5}\mathbf{W}_\phi),
\end{align}
where $\times_n$ denotes the $n$-mode tensor product, $\mathbf{W}_\phi\in\mathbb{R}^{C'\times UV}$ is the perceptron parameters, and the output is $f(\mathbf{S})\in\mathbb{R}^{T\times X\times Y\times L\times C'}$.
The MLP feature extraction can thus be formulated as
\begin{align}
    \mathbf{F} &= (f_n \circ f_{n-1} \circ \cdots \circ f_{1})(\mathbf{S}), 
\end{align} 
which produces a feature tensor $\mathbf{F}\in\mathbb{R}^{T\times X\times Y\times L\times C_F}$. 
This method is more flexible and may also be more effective than the soft-argmax because not only can it encode displacement information but also it can directly access the similarity values, which may be helpful for learning motion distribution.

\noindent \textbf{Convolution.}
The third method is to learn convolution kernels over $(L,U,V)$ volume of $\mathbf{S}$ (Fig.~\ref{fig:conv}).
When we regard $\mathbf{S}$ as a 7D tensor $\mathbf{S} \in \mathbb{R}^{T\times X\times Y\times L\times U\times V\times C}$ with $C=1$, the convolution layer $g(\cdot)$ can be expressed as
\begin{align}
    g(\mathbf{S}) &= \mathrm{ReLU} (\mathrm{Conv}(\mathbf{S}, \mathbf{K}_e)),
\end{align}
where $\mathbf{K}_e\in\mathbb{R}^{1\times 1\times 1\times L_{\kappa} \times U_{\kappa} \times V_{\kappa} \times C\times C'}$ is a multi-channel convolution kernel.
Starting from $\mathbb{R}^{T\times X\times Y\times L\times U\times V\times 1}$, we gradually downsample (U,V) and expand channels via multiple convolutions with strides, finally resulting in $\mathbb{R}^{T\times X\times Y\times L\times 1\times 1\times C_F}$; we preserve the $\textrm{L}$ dimension, since maintaining fine temporal resolution is shown to be effective for capturing detailed motion information~\cite{lin2019tsm, feichtenhofer2019slowfast}.
In practice, we reshape $\mathbf{S}$ and then apply a regular 3D convolution along $(l,u,v)$ dimension of $\mathbf{S}$.
The convolutional feature extraction with $n$ layers can thus be formulated as 
\begin{align}
    \mathbf{F} &= (g_n \circ g_{n-1} \circ \cdots \circ g_{1})(\mathbf{S}), 
\end{align}
which results in a feature tensor $\mathbf{F} \in \mathbb{R}^{T\times X\times Y\times L\times C_F}$. 
This method is effective in learning structural patterns with their convolution kernels, thus outperforming the former methods as will be seen in our experiments.

\subsubsection{Feature integration} \label{sec:selfy}
In this step, we integrate the extracted STSS features $\mathbf{F} \in \mathbb{R}^{T\times X\times Y\times L\times C_F}$ to feed them back to the original input stream with $(T, X, Y, C)$ volume.     

We first use spatio-temporal convolution kernels along $(t,x,y)$ dimension of $\mathbf{F}$.
The convolution layer $h(\cdot)$ can be expressed as
\begin{align}
    h(\mathbf{F}) &= \mathrm{ReLU} (\mathrm{Conv}(\mathbf{F}, \mathbf{K}_{i})),
\end{align}
where $\mathbf{K}_i\in\mathbb{R}^{T_{\kappa}\times X_{\kappa}\times Y_{\kappa}\times 1\times C_F\times C'_F}$ is a multi-channel convolution kernel.
This type of convolution integrates the extracted STSS features by extending receptive fields along $(t,x,y)$ dimension.
In practice, we reshape $\mathbf{F}$ and then apply a regular 3D convolution along $(t,x,y)$ dimension of $\mathbf{F}$.
The resultant features $\mathbf{F}^{\star} \in \mathbb{R}^{T\times X\times Y\times L\times C^{\star}_F}$ is defined as
\begin{align}
    \mathbf{F}^{\star} = (h_n \circ h_{n-1} \circ \cdots \circ h_{1})(\mathbf{F}).
\end{align}
We then flatten the $(L,C^{\star}_F)$ volume into $LC^{\star}_F$-dimensional vectors to obtain $\mathbf{F}^{\star} \in \mathbb{R}^{T\times X\times Y\times LC^{\star}_F}$, and apply an $1\times 1\times 1$ convolution layer to obtain the final output.
This convolution layer integrates features from different temporal offsets and also adjusts its channel dimension to fit that of the original input $\mathbf{V}$.
The final output tensor $\mathbf{Z}$ is expressed as
\begin{align}
\label{eq:Z}
    \mathbf{Z} = \mathrm{ReLU}(\mathbf{F}^{\star} \times_4 \mathbf{W_\theta}),      
\end{align}
where $\times_n$ is the $n$-mode tensor product and $\mathbf{W_\theta}\in\mathbb{R}^{C\times LC^{\star}_F}$ is the weights of the convolution layer.

Finally, we combine the resultant STSS representation
$\mathbf{Z}$ into the input feature $\mathbf{V}$ by element-wise addition, thus making SELFY act as a residual block~\cite{he2016deep}.

%% file: sections/4_experiment.tex
\section{Experiments}

\begin{table*}[t]

\centering
\scalebox{0.88}{
\begin{tabular}{lccccccc}
\toprule[1pt]
model & flow & \#frame  & FLOPs$\times$clips & \multicolumn{2}{c}{SS-V1} & \multicolumn{2}{c}{SS-V2}	\\
   &    &  &   & top-1 & top-5  & top-1 & top-5 \\
\midrule
TSN-R50 from~\cite{lin2019tsm} &		& 8 & 33 G$\times$1  & 19.7 & 46.6 & 30.0 & 60.5 \\
TRN-BNIncep~\cite{zhou2018temporal} &		& 8  & 16 G$\times$N/A  & 34.4 & - & 48.8 & - \\
TRN-BNIncep Two-Stream~\cite{zhou2018temporal} &	\cmark	& 8+8  & 16 G$\times$N/A  & 42.0 & - & 55.5 & - \\
MFNet-R50~\cite{lee2018motion} &		& 10  & N/A$\times$10 & 40.3 & 70.9 & - & - \\
CPNet-R34~\cite{liu2019learning} &		& 24  & N/A$\times$96  & - & - & 57.7 & 84.0 \\
TPN-R50~\cite{yang2020temporal} &	& 8   & N/A$\times$10   & 40.6 &   - & 59.1 & -    \\
\midrule
SELFYNet-R50 (ours) &	& 8 & 37 G$\times$1  & \textbf{50.7} & \textbf{79.3} & \textbf{62.7} & \textbf{88.0}  \\ 
\midrule
\midrule
I3D from~\cite{wang2018videos}& 	& 32 & 153 G$\times$2  & 41.6 & 72.2 & - & - \\
NL-I3D from~\cite{wang2018videos}& 	& 32  & 168 G$\times$2  & 44.4 & 76.0 & - & - \\
TSM-R50~\cite{lin2019tsm} & 	& 16    & 65 G$\times$1  & 47.3 &   77.1 & 61.2 & 86.9   \\
TSM-R50 Two-Stream from~\cite{kwon2020motionsqueeze} & \cmark   	& 16+16  & 129 G$\times$1 & 52.6 &   81.9 & 65.0 & 89.4\\
CorrNet-R101~\cite{wang2020video} &	& 32   & 187 G$\times$10  & 50.9 &   - & - & -    \\
STM-R50~\cite{jiang2019stm} &	& 16   & 67 G$\times$30  & 50.7 &   80.4 & 64.2 & 89.8    \\
TEA-R50~\cite{li2020tea} &	& 16   & 70 G$\times$30  & 52.3 &   81.9 & - & -    \\
MSNet-TSM-R50~\cite{kwon2020motionsqueeze} &	& 16 & 67 G$\times$1  & 52.1 & 82.3 & 64.7 & 89.4  \\  
MSNet-TSM-R50$_{EN}$~\cite{kwon2020motionsqueeze} &	 & 8+16 & 101 G$\times$10 & 55.1 & 84.0 & 67.1 & 91.0  \\
\midrule
SELFYNet-TSM-R50 (ours) &	& 8 & 37 G$\times$1 & 52.5 & 80.8 & 64.5 & 89.4  \\ 
SELFYNet-TSM-R50 (ours) &	& 16 & 77 G$\times$1  & 54.3 & 82.9 & 65.7 & 89.8  \\ 
SELFYNet-TSM-R50$_{EN}$ (ours) &	& 8+16 & 114 G$\times$1 & 55.8 & 83.9 & 67.4 & 91.0  \\ 
SELFYNet-TSM-R50$_{EN}$ (ours) &	& 8+16 & 114 G$\times$2 & \textbf{56.6} & \textbf{84.4} & \textbf{67.7} & \textbf{91.1}  \\ 
\bottomrule[1pt]
\end{tabular}
}
\captionsetup{width=0.85\linewidth}
\caption{\textbf{Performance comparison on SS-V1\&V2}. Top-1, 5 accuracy (\%) and FLOPs (G) are shown.
\label{sota_table}}
\end{table*}

\subsection{Implementation details} \label{sec:impl_details}

\noindent \textbf{Action recognition architecture.}
We employ TSN ResNets~\cite{wang2016temporal} as 2D CNN backbones and TSM ResNets~\cite{lin2019tsm} as 3D CNN backbones.
TSM enables to obtain the effect of spatio-temporal convolutions using spatial convolutions by shifting a part of input channels along the temporal axis before the convolution operation.
TSM is inserted into each residual block of the ResNet.
We adopt ImageNet pre-trained weights for our backbones.
To transform the backbones to the self-similarity network (SELFYNet), we insert a single SELFY block after the third stage in the backbone with additive fusion.
For the feature extraction and integration in SELFY block, we use four $1\times 3\times 3$ convolution layers along $(l,u,v)$ dimensions and four $1\times3\times 3$ convolution layers along $(t,x,y)$ dimensions, respectively.
For more details, please refer to supplementary material A.

\noindent \textbf{Training \& testing.}
For training, we sample a clip of 8 or 16 frames from each video using segment-based sampling~\cite{wang2016temporal}.
The spatio-temporal matching region $(L, U, V)$ of SELFY block is set as $(5, 9, 9)$ or $(9, 9, 9)$ when using 8 or 16 frames, respectively.
For testing, we sample one or two clips from a video, crop their center, and evaluate the averaged prediction of the sampled clips.
For more details, please refer to supplementary material A.

\subsection{Datasets}

For evaluation, we use benchmarks that contain fine-grained spatio-temporal dynamics in videos.

\noindent \textbf{Something-Something V1 \& V2 (SS-V1 \& V2)}~\cite{goyal2017something}, which are both large-scale action recognition datasets, contain $\sim$108k and $\sim$220k video clips, respectively. Both datasets share the same 174 action classes that are labeled, \eg, `pretending to put something next to something.'

\noindent \textbf{Diving-48}~\cite{li2018resound}, which contains $\sim$18k videos with 48 different diving action classes, is an action recognition dataset that minimizes contextual biases, \ie, scenes or objects.

\noindent \textbf{FineGym}~\cite{shao2020finegym} is a fine-grained action dataset built on top of gymnastic videos. We adopt the {\em Gym288} and {\em Gym99} sets that contain 288 and 99 classes, respectively.

\begin{table}[]
\centering
\scalebox{0.9}{
\begin{tabular}{lccc}
\toprule
model & \#frame & FLOPs & Top-1 \\
        &    & $\times$clips &  \\
\midrule
TSN from~\cite{li2018resound}            & -   & - & 16.8     \\
TRN from~\cite{kanojia2019attentive}            & -  & -  & 22.8     \\
Att-LSTM~\cite{kanojia2019attentive}            & 64 & N/A$\times$1 & 35.6     \\ 
P3D from~\cite{luo2019grouped}            & 16  & N/A$\times$1  & 32.4   \\
C3D from~\cite{luo2019grouped}            & 16 & N/A$\times$1   & 34.5    \\
GST-R50~\cite{luo2019grouped}       & 16  & 59 G$\times$1 &   38.8  \\
CorrNet-R101~\cite{wang2020video}   & 32  & 187 G$\times$10   &  38.2  \\ 
GSM-IncV3~\cite{sudhakaran2020gate}       & 16 & 54 G$\times$2    & 40.3  \\
\midrule
TSM-R50 (our impl.)                    & 16 & 65 G$\times$2   &   38.8  \\   
SELFYNet-TSM-R50 (ours)       & 16   & 77 G$\times$2  &  \textbf{41.6} \\            
\bottomrule
\end{tabular}
}
\caption{\textbf{Performance comparison on Diving-48}. Top-1 accuracy (\%) and FLOPs (G) are shown. \label{diving_table}}

\vspace{-2mm}

\end{table}

\begin{table}[]
\centering

\scalebox{0.9}{
\begin{tabular}{lccc}
\toprule
model & \#frame & Gym288 & Gym99 \\
        &       & Mean & Mean \\
\midrule
TSN~\cite{wang2016temporal}            & 3     & 26.5 & 61.4     \\
TRN~\cite{zhou2018temporal}             & 3     & 33.1 & 68.7     \\
I3D~\cite{carreira2017quo}            & 8     & 27.9 & 63.2   \\
NL I3D~\cite{wang2018non}            & 8     & 27.1 & 62.1   \\    
TSM~\cite{lin2019tsm}              & 3     & 34.8 & 70.6   \\
TSM Two-Stream~\cite{lin2019tsm}      & N/A & 46.5 & 81.2    \\
\midrule
TSM-R50 (our impl.)                    & 3     & 35.3  & 73.7  \\             
TSM-R50 (our impl.)                    & 8     & 47.9  & 86.6  \\   
SELFYNet-TSM-R50 (ours)       & 8  & \textbf{49.5}  & \textbf{87.7} \\            
\bottomrule
\end{tabular}
}
\caption{\textbf{Performance comparison on FineGym}. The averaged per-class accuracy (\%) is shown. All results in the upper part are from FineGym paper~\cite{shao2020finegym}. \label{finegym_table}}

\vspace{-4.2mm}

\end{table}

\subsection{Comparison with the state-of-the-art methods}

For a fair comparison, we compare our model with other models that are not pre-trained on additional large-scale video datasets, \eg, Kinetics~\cite{kay2017kinetics} or Sports1M~\cite{karpathy2014large}, in the following experiments.

Table~\ref{sota_table} summarizes the results on SS-V1\&V2.
The first and second compartment of the table shows the results of other 2D CNN and (pseudo-) 3D CNN models, respectively. 
The last part of each compartment shows the results of SELFYNet.
SELFYNet with TSN-ResNet (SELFYNet-TSN-R50) achieves 50.7\% and 62.7\% at top-1 accuracy, respectively, which outperforms other 2D models using 8 frames only.
When we adopt TSM ResNet (TSM-R50) as our backbone and use 16 frames, our method (SELFYNet-TSM-R50) achieves 54.3\% and 65.7\% at top-1 accuracy, respectively, which is the best among the single models. 
Compared to TSM-R50, a single SELFY block obtains the significant gains of 7.0\%p and 4.5\%p at top-1 accuracy, respectively; our method is more accurate than TSM-R50 two-stream on both datasets.
Finally, our ensemble model (SELFYNet-TSM-R50$_{EN}$) with 2-clip evaluation sets a new state-of-the-art on both datasets by achieving 56.6\% and 67.7\% at top-1 accuracy, respectively.

Tables~\ref{diving_table} and \ref{finegym_table} summarize the results on Diving-48 and FineGym.
For Diving-48, TSM-R50 using 16 frames shows 38.8\% at top-1 accuracy in our implementation.
SELFYNet-TSM-R50 outperforms TSM-R50 by 2.8\%p at top-1 accuracy so that it sets a new state-of-the-art top-1 accuracy as 41.6\% on Diving-48.
For FineGym, SELFYNet-TSM-R50 achieves 49.5\% and 87.7\% at given 288 and 99 classes, respectively, surpassing all the other models reported in~\cite{shao2020finegym}.

\subsection{Ablation studies} \label{sec:ablation}

We conduct ablation experiments to demonstrate the effectiveness of the proposed method.
All experiments are performed on SS-V1 using 8 frames.
Unless specified otherwise, we set ImageNet pre-trained TSM ResNet-18 (TSM-R18) with the single SELFY block of which $(L,U,V)=(5,9,9)$, as our default SELFYNet.

\noindent \textbf{Types of similarity.}
In Table~\ref{ablation_longterm}, we investigate the effect of different types of similarity by varying the set of temporal offset $l$ on both TSN-ResNet-18 (TSN-R18) and TSM-R18.
Interestingly, learning spatial self-similarity ($\{0\}$) improves accuracy on both backbones, which implies that self-similarity features help capture structural patterns of visual features.
Learning cross-similarity with a short temporal range  ($\{1\}$) shows a noticeable gain at accuracy on both backbones, indicating the significance of motion features.
Learning STSS outperforms other types of similarity, and the accuracy of SELFYNet increases as the temporal range becomes longer.
When STSS takes a far-sighted view on motion, STSS learns both short-term and long-term interactions in videos, as well as spatial self-similarity.

\noindent \textbf{Feature extraction and integration methods.} 
In Table~\ref{ablation_fe_fi}, we compare the performance of different combinations of feature extraction and integration methods.
From the 2$^\textrm{nd}$ to the 4$^\textrm{th}$ rows, different feature extraction methods are compared, fixing the feature integration methods to a single fully-connected (FC) layer.
Compared to the baseline, the use of soft-argmax, which extracts spatial displacement features, improves the top-1 accuracy by 1.0\%p. 
Replacing soft-argmax with MLP provides the additional gain of 1.9\%p at top-1 accuracy, showing the effectiveness of directly using similarity values.
When using the convolution method for feature extraction, we achieve 46.7\% at top-1 accuracy; the multi-channel convolution kernel is more effective in capturing structural patterns along $(u,v)$ dimensions than MLP. 
From the 4$^\textrm{th}$ to the 6$^\textrm{th}$ rows, different feature integration methods are compared, fixing the feature extraction method to convolution.
Replacing the single FC layer with MLP improves the top-1 accuracy by 0.6\%p.
Replacing MLP with convolutional layers further improves and achieves 48.4\% at top-1 accuracy. These results demonstrate that our design choice of using convolutions along $(u,v)$ and $(h,w)$ dimensions is the most effective in learning the geometry-aware STSS representation.
For more experiments, please refer to supplementary material C.

\subsection{Relation with self-attention mechanisms}
Note that self-similarity is also used in self-attention mechanisms~\cite{vaswani2017attention, hu2019local,ramachandran2019stand,srinivas2021bottleneck,wang2018non}, but both the purpose and the scheme are very different.
Self-attention mechanisms aim to perform dynamic feature transformation based on the image context and thus use the self-similarity as attention weights in aggregating individual features.
In contrast, our method focuses on learning relational representation from the self-similarity tensor itself.
We directly transform the tensor into a relational representation with learnable convolution kernels, where the relational representation of video is interpreted as generalized motion representation.

For an apple-to-apple empirical validation, we compare our method with popular self-attention methods~\cite{ramachandran2019stand,srinivas2021bottleneck,wang2018non}.
We re-implement the local self-attention~\cite{ramachandran2019stand} and Transformer~\cite{srinivas2021bottleneck} blocks, and extend them to a temporal dimension.
For a fair comparison, we insert a single block after $res_3$ of ResNet-18.
All other experimental details are the same as those in supplementary material A.
Table~\ref{ablation_NL} summarizes the results.
Our method outperforms the self-attention methods at both top-1 and top-5 accuracies with large margins.
These results demonstrate that learning the STSS representation effectively leverages motion features, which play a crucial role in action recognition.
For more experiments, please refer to supplementary material C.

\begin{table}[t]
    \centering
    \begin{subtable}[t]{\linewidth}
        \centering
        \small
        \begin{tabular}[t]{L{1.6cm}C{1.9cm}|C{0.9cm}C{0.8cm}C{0.8cm}}
            \toprule
            model & range of $l$    & FLOPs  & top-1 & top-5 \\
            \hline
            TSN-R18    & -   & 14.6 G  &16.2  &40.8 \\
            \hline
            &$\{0\}$         & 15.3 G  &16.8  &42.2 \\
            &$\{1\}$         & 15.3 G  &39.7  &68.9 \\ 
            SELFYNet & $\{-1,0,1\}$   & 16.3 G  &44.7  &73.9 \\
                     &$\{-2,\cdots,2\}$ & 17.3 G &\textbf{46.9}  &75.9 \\
                     &$\{-3,\cdots,3\}$ & 18.3 G &\textbf{46.9}  &\textbf{76.2} \\
            \midrule
            \midrule
            TSM-R18 & -  & 14.6 G  &43.0  &72.3 \\
            \hline
            &$\{0\}$ & 15.3 G &45.0  &73.4 \\
            &$\{1\}$ & 15.3 G &47.1  &76.3 \\
            SELFYNet & $\{-1,0,1\}$ & 16.3 G  &47.8  &76.7  \\
            &$\{-2,\cdots,2\}$ & 17.3 G &48.4  &77.6 \\
            &$\{-3,\cdots,3\}$ & 18.3 G &\textbf{48.6}  &\textbf{77.7} \\
            \bottomrule
        \end{tabular}
        \captionsetup{width=\linewidth}
        \caption{\textbf{Types of similarity}. Performance comparison with different sets of temporal offset in SELFY block. $\{\cdot\}$ denotes a set of temporal offset $l$.            \label{ablation_longterm}}     
    \end{subtable}   
    
    \begin{subtable}[t]{\columnwidth}
        \centering
        \small
            \begin{tabular}[t]{L{1.6cm}C{1.4cm}C{1.4cm}|C{0.8cm}C{0.8cm}} 	            
                \toprule
                model  & extraction & integration & top-1 & top-5 \\
                \hline
                TSM-R18                     & - & - & 43.0  & 72.3   \\
                \hline
                \multirow{5}{*}{SELFYNet} & Smax & FC & 44.0  & 72.3  \\ 
                & MLP  & FC & 45.9  & 75.1  \\ 
                & Conv & FC & 46.7  & 75.8  \\ 
                & Conv & MLP & 47.2  & 75.9  \\ 
                & Conv & Conv & \textbf{48.4}  &\textbf{77.6}  \\ 
                \bottomrule
            \end{tabular}
        \captionsetup{width=\columnwidth}
        \caption{\textbf{Feature extraction and integration methods.} Smax denotes the soft-argmax operation. MLP consist of four FC layers. The $1 \times 1 \times 1$ layer in the feature integration stage is omitted.
                      \label{ablation_fe_fi}  }             
    \end{subtable}
    \captionsetup{width=\linewidth}
    \caption{\textbf{Ablations on SS-V1}. Top-1 \& 5 accuracy (\%) are shown.}     
\end{table}

\begin{table}[t]
    \centering
    \small
        \begin{tabular}[t]{L{3.2cm}C{1.7cm}|C{0.8cm}C{0.8cm}} 	
            \toprule
            model  & $(L,U,V)$ & top-1 & top-5 \\
            \hline
            TSM-R18        & - & 43.0  &72.3   \\
            \hline
            TSM-R18 + LSA~\cite{ramachandran2019stand}   &$(5,9,9)$ & 43.8  & 72.8   \\
            TSM-R18 + NL~\cite{wang2018non}   & global & 43.5  & 73.4   \\
            TSM-R18 + MHSA~\cite{srinivas2021bottleneck}   & global & 44.0  & 72.8   \\
            \hline
            SELFYNet        &$(5,9,9)$ & \textbf{48.4}  &\textbf{77.6}  \\ 
            \bottomrule
        \end{tabular}  
        \captionsetup{width=\columnwidth}
        \caption{\textbf{Performance comparison with self-attention methods~\cite{ramachandran2019stand,srinivas2021bottleneck,wang2018non}.} LSA, NL, and MHSA denote a local self-attention block~\cite{ramachandran2019stand}, non-local block~\cite{wang2018non}, and multi-head self-attention block~\cite{srinivas2021bottleneck}, respectively.  \label{ablation_NL} }
\end{table}

\begin{figure*}[t]
    \begin{minipage}{0.52\linewidth}
    \centering
        \begin{minipage}{0.26\linewidth}
            \begin{subfigure}{\linewidth}
                \centering
                \includegraphics[width=\linewidth]{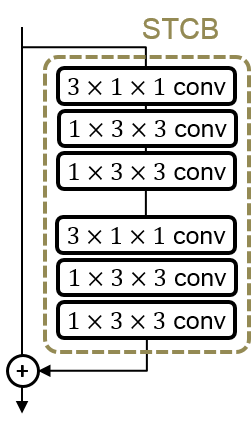}
                \vspace{-9mm}
                \subcaption{}
                \label{fig4:sub1}
            \end{subfigure}        
        \end{minipage}
        \begin{minipage}{0.29\linewidth}
            \begin{subfigure}{\linewidth}
                \centering
                \includegraphics[width=\linewidth]{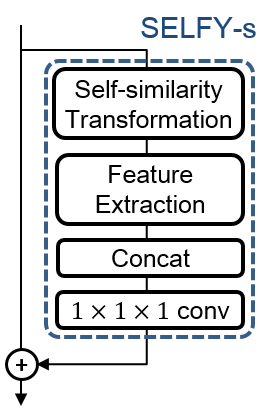}
                \vspace{-9mm}                
                \subcaption{}
            \label{fig4:sub2}
            \end{subfigure}        
        \end{minipage}
        \begin{minipage}{0.14\linewidth}
        \centering
            \begin{subfigure}{\linewidth}
                \centering
                \includegraphics[width=\linewidth]{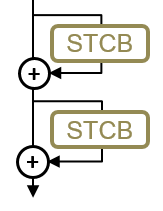} \\[1ex]
                \vspace{-4mm}                
                \subcaption{} 
            \label{fig4:sub3}
            \end{subfigure}  
            \begin{subfigure}{\linewidth}
                \centering
                \includegraphics[width=\linewidth]{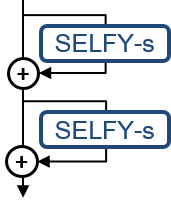}
                \vspace{-5.4mm}                
                \subcaption{}
            \label{fig4:sub4}
            \end{subfigure}                 
        \end{minipage}  
        \begin{minipage}{0.24\linewidth}
        \centering
            \begin{subfigure}{\linewidth}
                \centering
                \includegraphics[width=\linewidth]{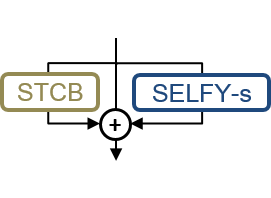} \\[1ex]
                \vspace{-3.5mm}                
                \subcaption{} 
            \label{fig4:sub5}
            \end{subfigure}  
            \begin{subfigure}{\linewidth}
                \centering
                \includegraphics[width=\linewidth]{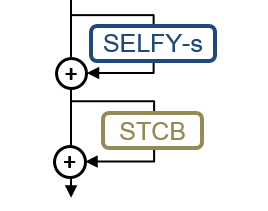}
                \vspace{-7mm}                
                \subcaption{}
            \label{fig4:sub6}
            \end{subfigure}  
        \end{minipage}   
        
    \captionsetup{width=\linewidth}    
    
    \vspace{-2mm}
     \caption{\textbf{Basic blocks and their combinations}. (a) spatio-temporal convolution block (STCB), (b) SELFY-s block, and (c-f) their different combinations.}
    \label{fig:4}
    \end{minipage}
\hfill
    \begin{minipage}{0.45\linewidth}
    \centering    
    \small
        \begin{tabular}[t]{L{4.9cm}|C{0.8cm}C{0.8cm}} 	
            \toprule
            model, TSN-R18      & top-1 & top-5 \\
            \hline
            baseline       &16.2  &40.8   \\
            \hline
            (a) STCB                           &42.4  &71.7  \\
            (b) SELFY-s                           &46.3  &75.1  \\                
            (c) STCB + STCB                          &44.4  &73.7  \\
            (d) SELFY-s + SELFY-s                          &46.8  &75.9  \\ 
            (e) SELFY-s + STCB (parallel)                       &46.9  &76.5  \\
            (f) SELFY-s + STCB (sequential)          &\textbf{47.6}  &\textbf{76.6}  \\
            \bottomrule
        \end{tabular}
        \captionsetup[table]{width=\linewidth}
        \captionof{table}{\textbf{Spatio-temporal features v.s. STSS features}. The basic blocks and their different combinations in Fig.~\ref{fig:4} are compared on SS-V1.         \label{complementary}}         
\end{minipage}

\vspace{-5mm}

\end{figure*}

\begin{figure*}[t]
\captionsetup[subfigure]{position=b}
\centering
\subcaptionbox{corruption: occlusion \label{fig:5a}}
{\includegraphics[width=.28\linewidth]{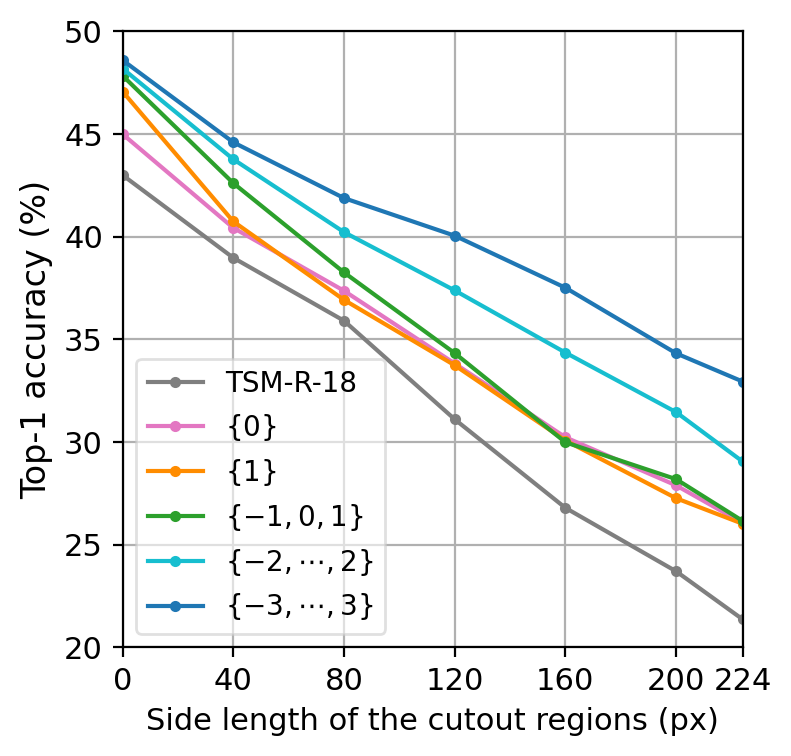}}
\hfill
\subcaptionbox{corruption: motion blur
\label{fig:5b}}{\includegraphics[width=.27\linewidth]{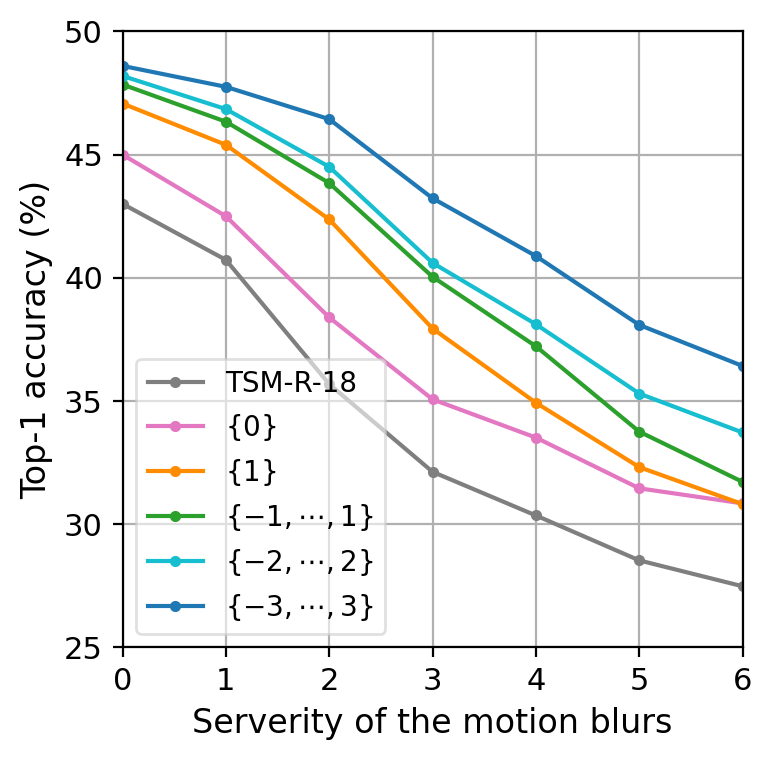}}
\hfill
\subcaptionbox{qualitative results on corrupted videos
\hfill
\label{fig:5c}}{\includegraphics[width=.42\linewidth]{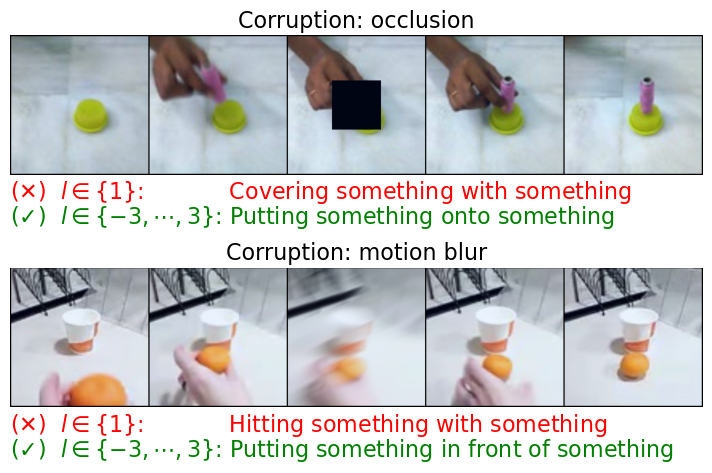}}
\captionsetup[figure]{position=b}
\vspace{-2mm}
\caption{\textbf{Robustness experiments}. (a) and (b) show top-1 accuracy of SELFYNet variants (Table~\ref{ablation_longterm}) when different degrees of occlusion and motion blur, respectively, are added to input. (c) shows qualitative examples where SELFYNet ($\{-3,\cdots,3\}$) succeeds while SELFYNet ($\{1\}$) fails.} \label{fig:5}

\vspace{-2mm}

\end{figure*}

\subsection{Complementarity of STSS features}

We conduct experiments for analyzing different meanings of spatio-temporal features and STSS features.
We organize two basic blocks for representing two different features: spatio-temporal convolution block (STCB) that consists of several spatial-temporal convolutions (Fig.~\ref{fig4:sub1}) and SELFY-s block, light-weighted version of the SELFY block by removing spatial convolution layers (Fig.~\ref{fig4:sub2}).
Both blocks have the same receptive fields and a similar number of parameters for a fair comparison.
Different combinations of the basic blocks are inserted after the third stage of TSN-ResNet-18.
Table~\ref{complementary} summarizes the results on SS-V1.
STSS features (Figs.~\ref{fig4:sub2} and \ref{fig4:sub4}) are more effective than spatio-temporal features (Figs.~\ref{fig4:sub1} and \ref{fig4:sub3}) at top-1 and top-5 accuracy when the same number of blocks are inserted.
Interestingly, the combination of two different features (Figs.~\ref{fig4:sub5} and \ref{fig4:sub6}) shows better results at top-1 and top-5 accuracy compared to the single feature cases (Figs.~\ref{fig4:sub3} and \ref{fig4:sub4}), which demonstrate that both features complement each other.
We conjecture that this complementarity comes from different characteristics of the two features;
while spatio-temporal features are obtained by directly encoding appearance features, STSS features are obtained by suppressing variations in appearance and focusing on the relational features in space and time.

\subsection{Improving robustness with STSS}

In this experiment, we demonstrate that STSS representation helps video-processing models to be more robust to video corruptions.
We test two types of corruption that are likely to occur in real-world videos: occlusion and motion blur.
To induce the corruptions, we either cut out a rectangle patch of a particular frame or generate a motion blur~\cite{hendrycks2019benchmarking}.
We corrupt a single center-frame for every clip of SS-V1 at the testing phase and gradually increase the severity of corruption.
We compare the results of TSM-R18 and SELFYNet variants of Table~\ref{ablation_longterm}.
Figures~\ref{fig:5a} and \ref{fig:5b} summarize the results of two corruptions, respectively.
Top-1 accuracy of TSM-R18 and SELFYNets with the short temporal range ($\{0\}$, $\{1\}$, and $\{-1,0,1\}$) significantly drops as the severity of corruption becomes harder.
We conjecture that features of the corrupted frame propagate through the stacked TSMs, confusing the entire network.
However, the SELFYNets with the long temporal range ($\{-2,\cdots,2\}$ and $\{-3,\cdots,3\}$) show more robust performance than the other models.
As shown in Figs.~\ref{fig:5a} and \ref{fig:5b}, the accuracy gap between SELFYNets with the long temporal range and the others increases as the severity of corruptions becomes higher, indicating that the larger size of STSS features can improve the robustness on action recognition.
We also present some qualitative results (Fig.~\ref{fig:5c}) where two SELFYNets with different temporal ranges, $\{1\}$ and $\{-3,\cdots,3\}$, both answer correctly without corruption, while the SELFYNet with $\{1\}$ fails for the corrupted input.

%% file: sections/5_conclusion.tex
\section{Conclusion}
We have proposed to learn a generalized, far-sighted motion representation from STSS for video understanding.
The comprehensive analyses on the STSS demonstrate that STSS features effectively capture both short-term and long-term interactions, complement spatio-temporal features, and improve the robustness of video-processing models.
Our method outperforms other state-of-the-art methods on the three benchmarks for video action recognition.

%% file: sections/6_supp.tex
\clearpage

\begin{center}
\textbf{\large Supplementary Material of ``Learning Self-Similarity in Space and Time as Generalized Motion for Video Action Recognition"}
\end{center}

\newcommand{\gray}[1]{\textcolor{gray}{#1}}

\setcounter{section}{0}
\renewcommand\thesection{\Alph{section}}

We present more experimental results that could not be included in the main manuscript due to the lack of space.

\section{Implementation details} \label{apdx:implementation_details}

\noindent \textbf{Architecture details.} We use TSN-ResNet and TSM-ResNet as our backbone (see Table~\ref{tab:backbone}) and initialize them with ImageNet pre-trained weights.
We insert a single SELFY block right after $res_3$ and use the convolution method as a default feature extraction method.
We set the spatio-temporal matching region of SELFY block, $(L,U,V)$, as $(5,9,9)$ or $(9,9,9)$ when using 8 or 16 input frames, respectively.
We stack four $1\times 3\times 3$ convolution layers along $(l,u,v)$ dimension for the feature extraction method, and use four $3\times 3$ convolution layers along $(x,y)$ dimension for the feature integration.
We reduce a spatial resolution of video feature tensor, $\mathbf{V}$, as 14$\times$14 for computation efficiency before the self-similarity transformation.
After the feature integration, we upsample the integrated feature tensor, $\mathbf{G^{\star}}$, as 28$\times$28 for the residual connection.

\noindent \textbf{Training.} We sample a clip of 8 or 16 frames from each video by using segment-based sampling~\cite{wang2016temporal}.
We resize the sampled clips into 240 $\times$ 320 images and apply random scaling and horizontal flipping for data augmentation.
When applying the horizontal flipping on SS-V1\&V2~\cite{goyal2017something}, we do not flip clips of which class labels include `left' or `right' words; the action labels, \eg, `pushing something from left to right.'
We fit the augmented clips into a spatial resolution of 224 $\times$ 224.
We adopt the SGD optimizer with a momentum of 0.9.
For SS-V1\&V2, we set the initial learning rate to 0.01 and the training epochs to 50; the learning rate is decayed by 1/10 after $30^\textrm{th}$ and $40^\textrm{th}$ epochs.
The training time of SELFYNet-TSM-R50 using 16 frames on SS-V1\&V2 is about 2$\sim$3 days with 8 Titan RTX GPUs.
For Diving-48~\cite{li2018resound} and FineGym~\cite{shao2020finegym}, we use a cosine learning rate schedule~\cite{loshchilov2016sgdr} with the first 10 epochs for gradual warm-up~\cite{goyal2017accurate}. 
We set the initial learning rate to 0.01 and the training epochs to 30 and 40, respectively.

\noindent \textbf{Testing.} Given a video, we sample 1 or 2 clips, resize them into 240 $\times$ 320 images, and crop their centers as 224 $\times$ 224.
We evaluate an average prediction of the sampled clips.
We report top-1 and top-5 accuracy for SS-V1\&V2 and Diving-48, and mean-class accuracy for FineGym.

\begin{table}[t]
\centering
\vspace{-5mm}
    \begin{center}
    \scalebox{0.8}{
    \begin{tabular}{c|c|c|c} 
    \hline
    Layers & TSN ResNet-50 & TSM ResNet-50 & Output size \\
    \hline 
    conv$_1$ & \multicolumn{2}{c|}{1$\times$7$\times$7, 64, stride 1,2,2}  & T$\times$112$\times$112 \\
    \hline
    pool$_1$ & \multicolumn{2}{c|}{1$\times$3$\times$3 max pool, stride 1,2,2} & T$\times$56$\times$56 \\
    \hline
    res$_2$ & $\begin{bmatrix}  $1$\times$1$\times$1, 256$ \\ $1$\times$3$\times$3, 256$ \\    $1$\times$1$\times$1, 256$ \end{bmatrix}\times$3  
    & $\begin{bmatrix}  $TSM$ \\ $1$\times$1$\times$1, 256$ \\ $1$\times$3$\times$3, 256$ \\    $1$\times$1$\times$1, 256$ \end{bmatrix}\times$3 & T$\times$56$\times$56 \\ 
    \hline
    res$_3$ & $\begin{bmatrix} $1$\times$1$\times$1, 512$ \\  $1$\times$3$\times$3, 512$ \\    $1$\times$1$\times$1, 512$ \end{bmatrix}\times$4 
    & $\begin{bmatrix} $TSM$ \\  $1$\times$1$\times$1, 512$ \\  $1$\times$3$\times$3, 512$ \\    $1$\times$1$\times$1, 512$ \end{bmatrix}\times$4 
    & T$\times$28$\times$28 \\
    \hline
    res$_4$ & $\begin{bmatrix} $1$\times$1$\times$1, 1024$ \\ $1$\times$3$\times$3, 1024$ \\    $1$\times$1$\times$1, 1024$ \end{bmatrix}\times$6 
    & $\begin{bmatrix} $TSM$\\  $1$\times$1$\times$1, 1024$ \\ $1$\times$3$\times$3, 1024$ \\    $1$\times$1$\times$1, 1024$ \end{bmatrix}\times$6 
    & T$\times$14$\times$14 \\
    \hline  
    res$_5$ & $\begin{bmatrix} $1$\times$1$\times$1, 2048$ \\ $1$\times$3$\times$3, 2048$ \\    $1$\times$1$\times$1, 2048$ \end{bmatrix}\times$3 
    & $\begin{bmatrix} $TSM$\\  $1$\times$1$\times$1, 2048$ \\ $1$\times$3$\times$3, 2048$ \\    $1$\times$1$\times$1, 2048$ \end{bmatrix}\times$3 
    & T$\times$7$\times$7 \\
    \hline  
    \multicolumn{3}{c|}{global average pool, FC} & \# of classes\\ 
      \hline
    \end{tabular}
    }
  \end{center}

\vspace{-3mm}

\caption{\textbf{TSN \& TSM ResNet-50 backbone.}     \label{tab:backbone} }  
\end{table}

\noindent \textbf{Frame corruption details.}
We adopt two corruptions, occlusion and motion blur, to test the robustness of SELFYNet.
We only corrupt a single center-frame for every validation clip of SS-V1; we corrupt the $4^\textrm{th}$ frame amongst 8 input frames.
For the occlusion, we cut out a rectangle region from the center of the frame.
For the motion blur, we adopt ImageNet-C implementation, which is available online\footnote{\url{https://github.com/hendrycks/robustness}}.
We set 6 levels of severity for each corruption.
We set the side length of the occluded region as 40px, 80px, 120px, 160px, 200px and 224px from the level 1 to 6.
For the motion blur, we set (\textit{radius}, \textit{sigma}) tuple arguments as $(15,5)$, $(10,8)$, $(15,12)$, $(20,15)$, $(25,20)$, and $(30,25)$.

\begin{table}[t]
\centering
\scalebox{0.76}{
    \begin{tabular}{lcccc}
    \toprule
    model & backbone &\#frames & FLOPs$\times$clips & top-1 \\
    \midrule
    STM~\cite{jiang2019stm}   & Res-50 & 16 & 67 G$\times$30  & 73.7   \\
    TSM~\cite{lin2019tsm}        & Res-50   & 16  & 65 G$\times$30 & 74.7     \\
    TEINet~\cite{liu2020teinet} & Res-50 & 16  & 66 G $\times$30   & 76.2     \\
    TEA~\cite{li2020tea}   &    Res-50     & 16 & 70 G$\times$30  & 76.1     \\ 
    MSNet-TSM~\cite{kwon2020motionsqueeze}    &   Res-50   & 16  & 67 G$\times$10  & 76.4   \\
    \midrule
    \gray{SlowFast $16\times8$+NL~\cite{feichtenhofer2019slowfast}}  &   \gray{3D Res-101}   & \gray{16+128}  & \gray{234 G$\times$30}  & \gray{79.8}   \\
    \gray{TimeSformer-L~\cite{bertasius2021space}}  &  \gray{ViT-L~\cite{dosovitskiy2020image}}    & \gray{96}  & \gray{2380 G$\times$3}  & \gray{80.7}   \\
    \midrule
    SELFYNet-TSM (ours)  & Res-50 & 16   & 77 G$\times$30  &  \textbf{77.1} \\            
    \bottomrule
    \end{tabular}
}
\small
\caption{\textbf{Performance comparison on Kinetics-400~\cite{kay2017kinetics}.} \label{kinetics_table}}
\end{table}

\section{Performance comparison on Kinetics-400}

We also conduct experiments on Kinetics-400~\cite{kay2017kinetics}, which is the most popular appearance-centric benchmark.
Table~\ref{kinetics_table} summarizes the results on Kinetics-400.
The first and second compartment of the table shows the results of different models with Res-50 using 16 frames and the results of the state-of-the-art models, respectively.
The last row shows our result.
The results demonstrate that SELFYNet still shows a clear improvement on the appearance-centric benchmark.
SELFYNet obtains the improvement of 2.4\%p at top-1 accuracy over the TSM baseline~\cite{lin2019tsm}, achieving the best accuracy among the models with Res-50 using 16 frames.
Although the accuracy of SELFYNet is inferior to that of SlowFast~\cite{feichtenhofer2019slowfast} or TimeSformer-L~\cite{bertasius2021space}, we expect that SELFYNet can achieve the state-of-the-art when using larger backbones (3D Res-101, ViT-L) or a bigger input.

In the following, we provide implementation details for Kinetics-400 experiments.
We adopt the dense frame sampling method~\cite{wang2018non} and sample a clip of 16 frames.
For training, we use a cosine learning rate schedule with the first 10 epochs for warm-up.
We set the initial learning rate to 0.01 and total epochs to 65.
For testing, we sample 10 uniform clips per video and average the softmax scores for the final prediction.
We follow the strategy of non-local networks~\cite{wang2018non} to pre-process the frames and take 3 crops as input.
Other experimental details are the same as those in the supplementary material~\ref{apdx:implementation_details}.

\begin{table*}[t]
    \centering
    \small
    \begin{subtable}[t]{0.93\columnwidth}
    \centering    
            \begin{tabular}[t]{L{1.6cm}C{1.6cm}|C{0.9cm}C{0.7cm}C{0.7cm}} 	
                \toprule
                model & $U\times V$    & FLOPs  & top-1 & top-5 \\
                \hline
                TSM-R18    & -         & 14.6 G  &43.0  &72.3   \\
                \hline
                \multirow{4}{*}{SELFYNet} & $5\times 5$ & 17.1 G   &47.8  &77.1  \\ 
                & $9\times 9$          & 17.3 G  &48.4  &77.6  \\
                & $13\times 13$           & 18.4 G  &48.4  &77.8  \\
                & $17\times 17$     & 19.8 G  &\textbf{48.6}  &\textbf{78.3}  \\
                \bottomrule
            \end{tabular}
        \captionsetup{width=\columnwidth}
            \caption{\textbf{Spatial matching region}. Performance comparison with different spatial matching-regions, ($U\times V$).  \label{ablation_region}}             
    \end{subtable}
    \qquad
    \begin{subtable}[t]{0.93\columnwidth}
        \centering
            \begin{tabular}[t]{L{2.3cm}C{2.2cm}|C{0.7cm}C{0.7cm}} 	
                \toprule
                model & position       & top-1 & top-5 \\
                \hline
                TSM-R18   & -          &43.0  &72.3   \\
                \hline
                \multirow{6}{*}{SELFYNet} &$\textrm{pool}_1$  &45.7  &77.6  \\ 
                &$\textrm{res}_2$             &47.2  &76.6  \\ 
                &$\textrm{res}_3$             &48.4  &77.6  \\
                &$\textrm{res}_4$             &46.6  &76.0  \\
                &$\textrm{res}_5$             &42.8  &72.6  \\ 
                &$\textrm{res}_{2,3,4}$       &\textbf{48.6}  &\textbf{77.9}  \\
                \bottomrule
            \end{tabular}
        \captionsetup{width=\columnwidth}
            \caption{\textbf{Position}. Performance comparison with different positions of SELFY block. For the last row, 3 SELFY blocks are used in total.             \label{ablation_position}}             
    \end{subtable}    

    \vspace{3mm}

    \begin{subtable}[t]{0.93\columnwidth}
    \centering
            \begin{tabular}[t]{L{2.3cm}C{2.2cm}|C{0.7cm}C{0.7cm}} 	
                \toprule
                model  & features  & top-1 & top-5 \\
                \hline
                TSM-R18                     & $\mathbf{V}$ & 43.0  &72.3   \\
                \hline
                \multirow{2}{*}{SELFYNet}   & $\mathbf{Z}$ & 45.5  & 75.9  \\ 
                & $\mathbf{Z+V}$ & \textbf{48.4}  &\textbf{77.6}  \\ 
                \bottomrule
            \end{tabular}        
        \captionsetup{width=\columnwidth}
            \caption{\textbf{STSS features with visual features.} $\mathbf{V},\mathbf{Z}$ denotes the visual features and STSS features, respectively. \label{ablation_Z}}       
    \end{subtable}
    \qquad
    \begin{subtable}[t]{0.93\columnwidth}
    \centering
            \begin{tabular}[t]{L{2.3cm}C{2.2cm}|C{0.7cm}C{0.7cm}} 	
                \toprule
                model & range of $l$       & top-1 & top-5 \\
                \hline
                TSM-R18    & -          &43.0  &72.3   \\
                \hline
                & $\{-1,0,1\}$  &47.4  &77.0  \\ 
                SELFYNet & $\{-2,\cdots,2\}$             &48.3  &77.2  \\
                & $\{-3,\cdots,3\}$             &\textbf{48.5}  &\textbf{77.4}  \\
                \bottomrule
            \end{tabular}
        \captionsetup{width=\columnwidth}
            \caption{\textbf{Multi-channel $3\times 3\times 3$ kernel for feature extraction.} Four convolution layers are used for extracting STSS features. $\{\cdot\}$ denotes a set of temporal offsets $l$. \label{ablation_3d}}
    \end{subtable}
    
    \vspace{3mm}

    \begin{subtable}[t]{0.93\columnwidth}
        \centering
        \begin{tabular}{L{1.45cm}C{1.3cm}C{1.45cm}|C{0.7cm}C{0.7cm}}
        \toprule
        model & similarity & extraction & top-1 & top-5 \\
        \hline
        TSM-R18    & -     & -     &43.0  &72.3   \\
        \hline
        \multirow{3}{*}{SELFYNet} & embed. G  & mult. w/ $\mathbf{V}$ & 43.8  &72.3  \\
        & embed. G  & Conv  & 47.6  &76.8  \\
        & cosine    & Conv  & 47.8  &77.1  \\
        \bottomrule
        \end{tabular}
        \captionsetup{width=\columnwidth}
        \caption{\textbf{Performance comparison with the local self-attention mechanisms~\cite{hu2019local,ramachandran2019stand}.} We implemented the local self-attention by following Ramachandran \etal~\cite{ramachandran2019stand}. 
        \label{ablation_lsa}} 
    \end{subtable}    
    \qquad    
    \begin{subtable}[t]{0.93\columnwidth}
        \vspace{-12mm}
        \centering
        \small
            \begin{tabular}[t]{L{1.6cm}C{1.3cm}C{1.3cm}|C{0.7cm}C{0.7cm}} 	
                \toprule
                model  & extraction & $(L,U,V)$ & top-1 & top-5 \\
                \hline
                TSM-R18                     & - & - & 43.0  &72.3   \\
                \hline
                \multirow{4}{*}{SELFYNet}   & KS + CM & $(1,9,9)$ & 46.1  & 75.3  \\ 
                & KS + CM  & $(5,9,9)$ & 47.4  & 76.8  \\ 
                & Conv & $(1,9,9)$ & 47.1  & 76.3  \\ 
                & Conv & $(5,9,9)$ & \textbf{48.4}  &\textbf{77.6}  \\ 
                \bottomrule
            \end{tabular}  
        \captionsetup{width=\columnwidth}
        \caption{\textbf{Performance comparison with MSNet~\cite{kwon2020motionsqueeze}.} KS and CM denote the kernel soft-argmax and confidence map, respectively. \label{ablation_MS}}
    \end{subtable}

        \begin{subtable}[t]{0.64\linewidth}
        \vspace{3mm}
        \centering
        \small
            
            \begin{tabular}[t]{lcccccc}             
                \toprule
                model  & frames & FLOPs & memory & runtime & top-1 & top-5 \\
                \hline
                TSM-R50~\cite{lin2019tsm}       & 8 & 33.1 G & 8.2 GB & 15.6 ms & 45.6  & 74.2   \\
                TSM-R50~\cite{lin2019tsm}       & 16 & 66.3 G & 15.7 GB & 30.1 ms & 47.3  & 77.1   \\
                \hline
                TSM-R50 + NL~\cite{wang2018non}   & 8 & 46.5 G & 10.3 GB & 24.0 ms & 49.1  & 77.2   \\
                TSM-R50 + MHSA~\cite{srinivas2021bottleneck}   & 8 & 50.6 G  & 15.9 GB & 26.3 ms & 49.2  & 77.9   \\
                \hline
                TSM-R50 + SELFY       & 8  & 36.6 G & 9.6 GB & 21.1 ms & \textbf{52.5}  & \textbf{80.8}  \\
                \bottomrule
            \end{tabular}  
            \captionsetup{width=\columnwidth}
        \small
            \caption{\textbf{Efficiency}. Performance comparison with other attention mechanisms~\cite{srinivas2021bottleneck,wang2018non}. We insert a single block after $\mathrm{res}_3$ in TSM-R50. We use 8 clips per GPU and measure the runtime by following protocols in \cite{kwon2020motionsqueeze}. \label{ablation_self_attention} }
    \end{subtable}

    \caption{\textbf{Additional experiments on SS-V1}. Top-1 \& 5 accuracy (\%) are shown.             \label{ablation_apdx}}     
\end{table*}

\section{Additional experiments} \label{apdx:additional_ablations}
We conduct additional experiments to identify the behaviors of the proposed method.
All experiments are performed on SS-V1 by using 8 frames.
Unless otherwise specified, we set ImageNet pre-trained TSM ResNet-18 (TSM-R18) with a single SELFY block of which $(L,U,V)=(5,9,9)$, as our default SELFYNet.

\noindent \textbf{Spatial matching region.}
In Table~\ref{ablation_region}, we compare a single SELFY block with different spatial matching regions, $(U,V)$.
As a result, indeed, the larger spatial matching region leads the better accuracy.
Considering the accuracy-computation trade-off, we set our spatial matching region, $(U,V)$, as $(9,9)$ as a default.

\noindent \textbf{Block position.} 
From the 2$^{\textrm{nd}}$ to the 6$^{\textrm{th}}$ row of Table~\ref{ablation_position}, we identify the effect of different positions of SELFY block in the backbone.
We resize the spatial resolution of the video tensor, $(X,Y)$, into 14$\times$14, and fix the matching region, $(L,U,V)$, as $(5,9,9)$ for all the cases maintaining the similar computational cost.
SELFY after the $\textrm{res}_3$ shows the best trade-off by achieving the highest accuracy among the cases; early-stage features ($\textrm{pool}_1$,$\textrm{res}_2$) lack enough semantics for robust matching while late-stage ones ($\textrm{res}_4$,$\textrm{res}_5$) lose appearance details for accurate matching.
The last row in Table~\ref{ablation_position} shows that the multiple SELFY blocks improve accuracy compared to the single block.

\noindent \textbf{Fusing STSS features with visual features.}
We evaluate SELFYNet purely based on STSS features to see how much the ordinary visual feature $\mathbf{V}$ contributes to the final prediction.
That is, we pass the STSS features, $\mathbf{Z}=\mathrm{ReLU}(\mathbf{F}^\star \times_5 \mathbf{W}_\theta)$, into the downstream layers without additive fusion.
Table~\ref{ablation_Z} compares the results of using different cases of the output tensor ($\mathbf{V}$, $\mathbf{Z}$, and $\mathbf{Z}+\mathbf{V}$) on SS-V1.
Interestingly, SELFYNet using only $\mathbf{Z}$ achieves 45.5\% at top-1 accuracy, which is higher as 2.5\%p than the baseline.
As we add $\mathbf{V}$ to $\mathbf{Z}$, we obtain an additional gain of 2.9\%p.
It indicates that the STSS features and the visual features are complementary to each other.

\noindent \textbf{Multi-channel $3\times 3\times 3$ kernel for feature extraction.}
We investigate the effect of the convolution method for STSS feature extraction when we use multi-channel $3\times 3\times 3$ kernels.
For the experiment, we stack four $3\times 3\times 3$ convolution layers followed by the feature integration step, which are the same as in Section 3.2.2 in our main manuscript.
Table~\ref{ablation_3d} summarizes the results.
Note that we do not report models of which temporal window $L=1$, \eg, $\{0\}$ and $\{1\}$.
As shown in the table, indeed, the long temporal range gives a higher accuracy.
However, the effect of the $3\times 3\times 3$ kernel is comparable to that of the $1\times 3\times 3$ kernel in Table 4a in our main manuscript.
Considering the accuracy-computation trade-off, we choose to fix the kernel size, $L_{\kappa} \times U_{\kappa} \times V_{\kappa}$, as $1\times 3\times 3$ for the STSS feature extraction.


\noindent \textbf{Relation with local self-attention mechanisms.}
The local self-attention~\cite{hu2019local,ramachandran2019stand,zhao2020exploring} and our method have a common denominator of using the self-similarity tensor but use it in a very different way and purpose.
The local self-attention mechanism aims to aggregate the local context features using the self-similarity tensor, and it thus uses the self-similarity values as attention weights for feature aggregation.
However, our method aims to learn a generalized motion representation from the local STSS, so the final STSS representation is directly fed into the neural network instead of multiplying it to local context features.

For an empirical comparison, we conduct an ablation experiment as follows. We extend the local self attention layer~\cite{ramachandran2019stand} to the temporal dimension and then add the \textit{spatio-temporal} local self-attention layer, which is followed by feature integration layers, after $res_3$. All experimental details are the same as those in supplementary material~\ref{apdx:implementation_details}, except that we reduce the channel dimension $C$ of appearance feature $\mathbf{V}$ to 32. Table~\ref{ablation_lsa} summarizes the results on SS-V1. The spatio-temporal local self-attention layer is accurate as 43.8\% at top-1 accuracy, and both of SELFY blocks using the embedded Gaussian and the cosine similarity outperform the local self-attention by achieving top-1 accuracy as 47.6\% and 47.8\%, respectively. These results are in alignment with the prior work~\cite{liu2019learning}, which reveals that the self-attention mechanism hardly captures motion in the video.

\noindent \textbf{Comparison with correlation-based methods.}
We also investigate the difference between our method and correlation-based methods~\cite{kwon2020motionsqueeze,wang2020video}.
While correlation-based methods extract motion features only from the spatial cross-similarity tensor between two adjacent frames, and are thus limited to short-term motion, our method effectively captures bi-directional and long-term motion information via learning with the sufficient volume of STSS.
Our method can also exploit richer information from the self-similarity values than other methods.
MS module~\cite{kwon2020motionsqueeze} only focuses on the maximal similarity value of the $(u,v)$ dimensions to extract flow information, and Correlation block~\cite{wang2020video} uses an $1\times1$ convolution layer for extracting motion features from the similarity values.
In contrast to the two methods, we introduce a generalized motion learning framework using the self-similarity tensor in Section 3.2 in our main manuscript.

We also conduct experiments to compare our method with MSNet~\cite{kwon2020motionsqueeze}, one of the correlation-based methods.
For an apple-to-apple comparison, we apply kernel soft-argmax and max pooling operation (\textit{KS} + \textit{CM} in~\cite{kwon2020motionsqueeze}) to our feature extraction method by following their official codes\footnote{\url{https://github.com/arunos728/MotionSqueeze}}.
Please note that, when we restrict the temporal offset $l$ to $\{ 1 \}$, the SELFY block using KS + CM is equivalent to the MS module of which \textit{feature transformation} layers are the standard 2D convolution layers.
Table~\ref{ablation_MS} summarizes the results.
KS+CM method achieves 46.1\% at top-1 accuracy.
As we enlarge the temporal window $L$ to 5, we obtain the additional gain as 1.3\%p.
The learnable convolution layers improve the top-1 accuracy by 1.0\%p in both cases.
The results demonstrate the effectiveness of learning geometric patterns within the sufficient
volume of STSS tensors for learning motion features.

\noindent \textbf{Efficiency.}
In Table~\ref{ablation_self_attention}, we compare the efficiency of SELFYNet with that of other self-attention methods~\cite{wang2018non,srinivas2021bottleneck} in terms of FLOPs, memory footprint, runtime, and accuracy.
Compared to TSM-R50 using 16 frames, SELFYNet using 8 frames consumes less memory by 6.1 GB and runs faster by 9.0 ms while improving top-1 accuracy by 5.2 \%p.
Compared to the self-attention methods~\cite{wang2018non,srinivas2021bottleneck}, SELFYNet also achieves the best accuracy with less memory footprint and faster inference speed.

\section{Visualizations} \label{apdx:visualizations}
\begin{figure*}[t]
    \centering
    \begin{subfigure}[t]{0.8\linewidth}
    \includegraphics[width=\columnwidth]{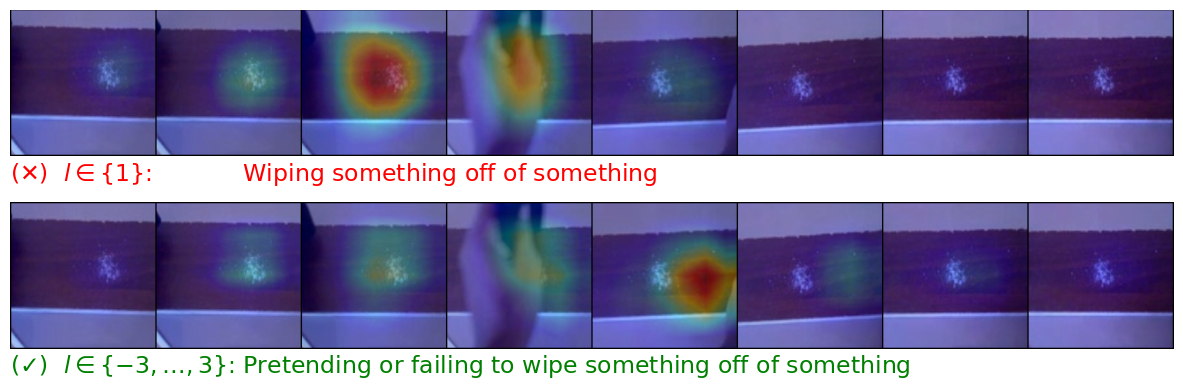}
    \vspace{-6mm}
    \caption{}
    \label{fig:s1a}
    \end{subfigure}
    \vspace{2mm}
    \begin{subfigure}[t]{0.8\linewidth}
    \includegraphics[width=\columnwidth]{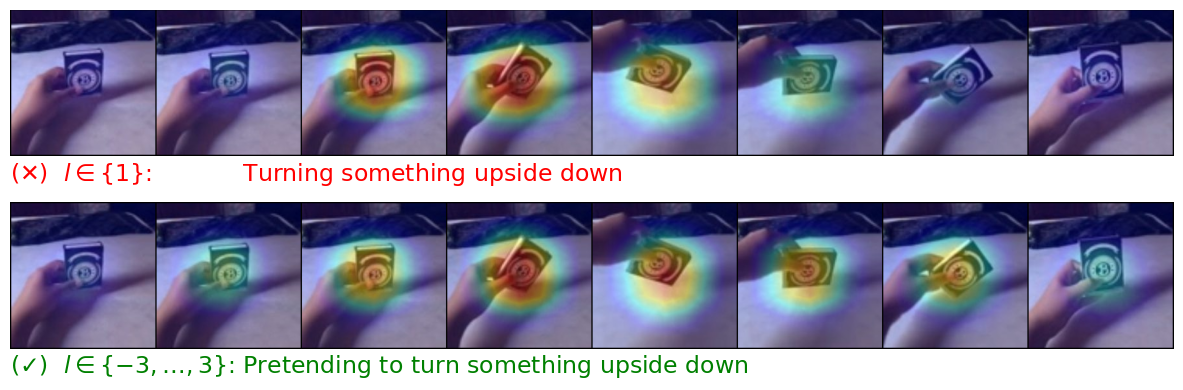}
    \vspace{-6mm}
    \caption{}
    \label{fig:s1b}
    \end{subfigure}
    \vspace{2mm}
    \begin{subfigure}[t]{0.8\linewidth}
    \includegraphics[width=\columnwidth]{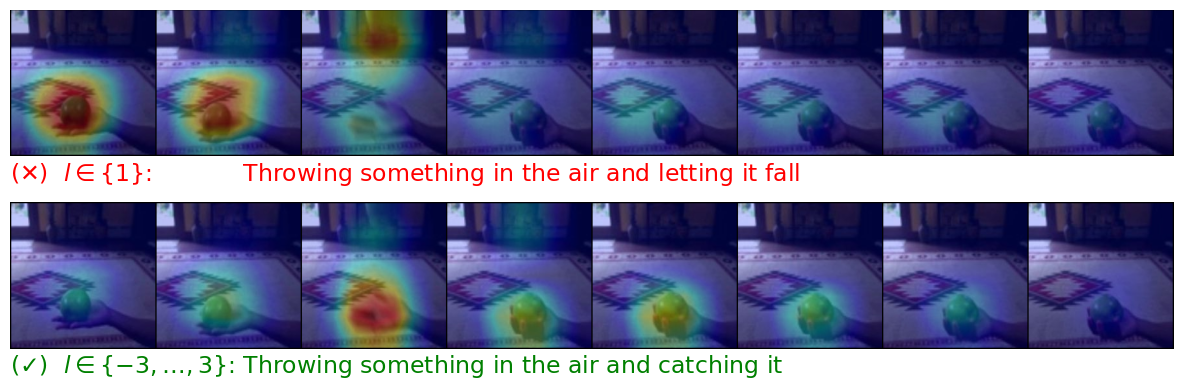}
    \vspace{-6mm}
    \caption{}
    \label{fig:s1c}
    \end{subfigure}
    \vspace{2mm}
    \begin{subfigure}[t]{0.8\linewidth}
    \includegraphics[width=\columnwidth]{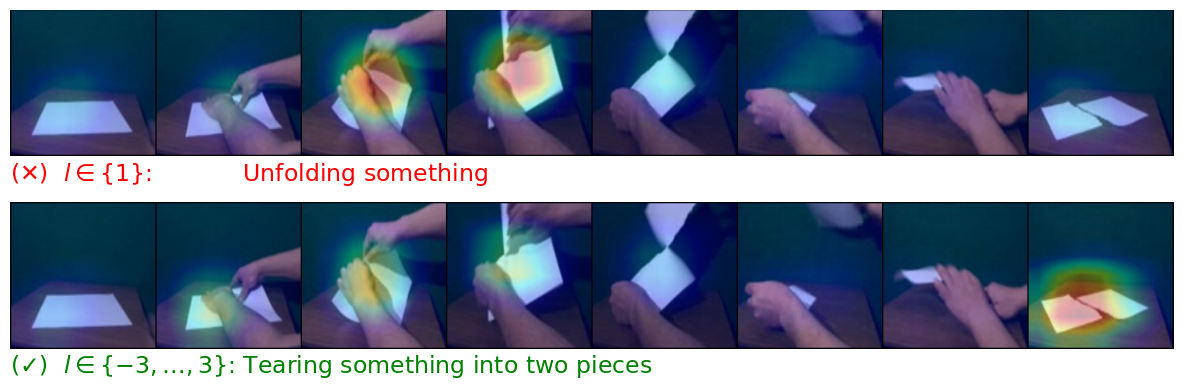}
    \vspace{-6mm}
    \caption{}
    \label{fig:s1d}
    \end{subfigure}
    
\caption{\textbf{Qualitative results} of two SELFYNets on SS-V1. Each subfigure visualizes prediction results of the two models with Grad-CAM-overlaid RGB frames.
The correct and wrong predictions are colorized as \textcolor{green}{green} and \red{red}, respectively.} \label{fig:s1}
\end{figure*}

In Fig.~\ref{fig:s1}, we visualize some qualitative results of two different SELFYNet-TSM-R18 ($\{1\}$ and $\{-3,\cdots,3\}$) on SS-V1.
We show the different predictions of the two models with 8 input frames.
We also overlay Grad-CAMs~\cite{selvaraju2017grad} on the input frames to see whether a larger volume of STSS benefits to capture long-term interactions in videos.
We take Grad-CAMs of features which is right before a global average pooling layer.
As shown in the figure, the STSS with the sufficient volume helps to learn the more enriched context of temporal dynamics in the video; in Fig.~\ref{fig:s1a}, for example, SELFYNet with the range of ($\{-3,\cdots,3\}$) focuses on not only regions on which an action occurs but also focuses on the white-stain after the action to verify whether the stain is wiped off or not.

%% file: main_arxiv.bbl
\begin{thebibliography}{10}\itemsep=-1pt

\bibitem{carreira2017quo}
Joao Carreira and Andrew Zisserman.
\newblock Quo vadis, action recognition? a new model and the kinetics dataset.
\newblock In {\em Proc. IEEE Conference on Computer Vision and Pattern
  Recognition (CVPR)}, 2017.

\bibitem{chapelle2010gradient}
Olivier Chapelle and Mingrui Wu.
\newblock Gradient descent optimization of smoothed information retrieval
  metrics.
\newblock {\em Information retrieval}, 13(3):216--235, 2010.

\bibitem{donahue2015long}
Jeffrey Donahue, Lisa Anne~Hendricks, Sergio Guadarrama, Marcus Rohrbach,
  Subhashini Venugopalan, Kate Saenko, and Trevor Darrell.
\newblock Long-term recurrent convolutional networks for visual recognition and
  description.
\newblock In {\em Proc. IEEE Conference on Computer Vision and Pattern
  Recognition (CVPR)}, 2015.

\bibitem{dosovitskiy2015flownet}
Alexey Dosovitskiy, Philipp Fischer, Eddy Ilg, Philip Hausser, Caner Hazirbas,
  Vladimir Golkov, Patrick Van Der~Smagt, Daniel Cremers, and Thomas Brox.
\newblock Flownet: Learning optical flow with convolutional networks.
\newblock In {\em Proc. IEEE International Conference on Computer Vision
  (ICCV)}, 2015.

\bibitem{bertasius2021space}
Bertasius et al.
\newblock Is space-time attention all you need for video understanding?
\newblock {\em arXiv preprint arXiv:2102.05095}, 2021.

\bibitem{dosovitskiy2020image}
Dosovitskiy et al.
\newblock An image is worth 16x16 words: Transformers for image recognition at
  scale.
\newblock {\em Proc. International Conference on Learning Representations
  (ICLR)}, 2021.

\bibitem{fan2020rubiksnet}
Linxi Fan, Shyamal Buch, Guanzhi Wang, Ryan Cao, Yuke Zhu, Juan~Carlos Niebles,
  and Li Fei-Fei.
\newblock Rubiksnet: Learnable 3d-shift for efficient video action recognition.
\newblock In {\em Proc. European Conference on Computer Vision (ECCV)}, 2020.

\bibitem{fan2018end}
Lijie Fan, Wenbing Huang, Chuang Gan, Stefano Ermon, Boqing Gong, and Junzhou
  Huang.
\newblock End-to-end learning of motion representation for video understanding.
\newblock In {\em Proc. IEEE Conference on Computer Vision and Pattern
  Recognition (CVPR)}, 2018.

\bibitem{feichtenhofer2020x3d}
Christoph Feichtenhofer.
\newblock X3d: Expanding architectures for efficient video recognition.
\newblock In {\em Proc. IEEE Conference on Computer Vision and Pattern
  Recognition (CVPR)}, 2020.

\bibitem{feichtenhofer2019slowfast}
Christoph Feichtenhofer, Haoqi Fan, Jitendra Malik, and Kaiming He.
\newblock Slowfast networks for video recognition.
\newblock In {\em Proc. IEEE International Conference on Computer Vision
  (ICCV)}, 2019.

\bibitem{girdhar2017attentional}
Rohit Girdhar and Deva Ramanan.
\newblock Attentional pooling for action recognition.
\newblock {\em arXiv preprint arXiv:1711.01467}, 2017.

\bibitem{goyal2017accurate}
Priya Goyal, Piotr Doll{\'a}r, Ross Girshick, Pieter Noordhuis, Lukasz
  Wesolowski, Aapo Kyrola, Andrew Tulloch, Yangqing Jia, and Kaiming He.
\newblock Accurate, large minibatch sgd: Training imagenet in 1 hour.
\newblock {\em arXiv preprint arXiv:1706.02677}, 2017.

\bibitem{goyal2017something}
Raghav Goyal, Samira~Ebrahimi Kahou, Vincent Michalski, Joanna Materzynska,
  Susanne Westphal, Heuna Kim, Valentin Haenel, Ingo Fruend, Peter Yianilos,
  Moritz Mueller-Freitag, et~al.
\newblock The" something something" video database for learning and evaluating
  visual common sense.
\newblock In {\em Proc. IEEE International Conference on Computer Vision
  (ICCV)}, 2017.

\bibitem{he2016deep}
Kaiming He, Xiangyu Zhang, Shaoqing Ren, and Jian Sun.
\newblock Deep residual learning for image recognition.
\newblock In {\em Proc. IEEE Conference on Computer Vision and Pattern
  Recognition (CVPR)}, 2016.

\bibitem{hendrycks2019benchmarking}
Dan Hendrycks and Thomas Dietterich.
\newblock Benchmarking neural network robustness to common corruptions and
  perturbations.
\newblock {\em arXiv preprint arXiv:1903.12261}, 2019.

\bibitem{hu2019local}
Han Hu, Zheng Zhang, Zhenda Xie, and Stephen Lin.
\newblock Local relation networks for image recognition.
\newblock In {\em Proceedings of the IEEE International Conference on Computer
  Vision}, pages 3464--3473, 2019.

\bibitem{jiang2019stm}
Boyuan Jiang, Mengmeng Wang, Weihao Gan, Wei Wu, and Junjie Yan.
\newblock Stm: Spatiotemporal and motion encoding for action recognition.
\newblock In {\em Proc. IEEE International Conference on Computer Vision
  (ICCV)}, 2019.

\bibitem{junejo2010view}
Imran~N Junejo, Emilie Dexter, Ivan Laptev, and Patrick Perez.
\newblock View-independent action recognition from temporal self-similarities.
\newblock {\em IEEE Transactions on Pattern Analysis and Machine Intelligence
  (TPAMI)}, 2010.

\bibitem{junejo2008cross}
Imran~N Junejo, Emilie Dexter, Ivan Laptev, and Patrick P{\'U}rez.
\newblock Cross-view action recognition from temporal self-similarities.
\newblock In {\em Proc. European Conference on Computer Vision (ECCV)}, 2008.

\bibitem{kang2021relational}
Dahyun Kang, Heeseung Kwon, Juhong Min, and Minsu Cho.
\newblock Relational embedding for few-shot classification.
\newblock In {\em Proc. IEEE International Conference on Computer Vision
  (ICCV)}, 2021.

\bibitem{kanojia2019attentive}
Gagan Kanojia, Sudhakar Kumawat, and Shanmuganathan Raman.
\newblock Attentive spatio-temporal representation learning for diving
  classification.
\newblock In {\em Proc. IEEE Conference on Computer Vision and Pattern
  Recognition Workshops (CVPRW)}, 2019.

\bibitem{karpathy2014large}
Andrej Karpathy, George Toderici, Sanketh Shetty, Thomas Leung, Rahul
  Sukthankar, and Li Fei-Fei.
\newblock Large-scale video classification with convolutional neural networks.
\newblock In {\em Proc. IEEE Conference on Computer Vision and Pattern
  Recognition (CVPR)}, 2014.

\bibitem{kay2017kinetics}
Will Kay, Joao Carreira, Karen Simonyan, Brian Zhang, Chloe Hillier, Sudheendra
  Vijayanarasimhan, Fabio Viola, Tim Green, Trevor Back, Paul Natsev, et~al.
\newblock The kinetics human action video dataset.
\newblock {\em arXiv preprint arXiv:1705.06950}, 2017.

\bibitem{kim2017fcss}
Seungryong Kim, Dongbo Min, Bumsub Ham, Sangryul Jeon, Stephen Lin, and
  Kwanghoon Sohn.
\newblock Fcss: Fully convolutional self-similarity for dense semantic
  correspondence.
\newblock In {\em Proc. IEEE Conference on Computer Vision and Pattern
  Recognition (CVPR)}, 2017.

\bibitem{kwon2020motionsqueeze}
Heeseung Kwon, Manjin Kim, Suha Kwak, and Minsu Cho.
\newblock Motionsqueeze: Neural motion feature learning for video
  understanding.
\newblock {\em arXiv preprint arXiv:2007.09933}, 2020.

\bibitem{kwon2018first}
Heeseung Kwon, Yeonho Kim, Jin~S Lee, and Minsu Cho.
\newblock First person action recognition via two-stream convnet with long-term
  fusion pooling.
\newblock {\em Pattern Recognition Letters}, 112:161--167, 2018.

\bibitem{lee2018motion}
Myunggi Lee, Seungeui Lee, Sungjoon Son, Gyutae Park, and Nojun Kwak.
\newblock Motion feature network: Fixed motion filter for action recognition.
\newblock In {\em Proc. European Conference on Computer Vision (ECCV)}, 2018.

\bibitem{li2020spatio}
Jun Li, Xianglong Liu, Mingyuan Zhang, and Deqing Wang.
\newblock Spatio-temporal deformable 3d convnets with attention for action
  recognition.
\newblock {\em Pattern Recognition}, 98:107037, 2020.

\bibitem{li2020smallbignet}
Xianhang Li, Yali Wang, Zhipeng Zhou, and Yu Qiao.
\newblock Smallbignet: Integrating core and contextual views for video
  classification.
\newblock In {\em Proc. IEEE Conference on Computer Vision and Pattern
  Recognition (CVPR)}, 2020.

\bibitem{li2020tea}
Yan Li, Bin Ji, Xintian Shi, Jianguo Zhang, Bin Kang, and Limin Wang.
\newblock Tea: Temporal excitation and aggregation for action recognition.
\newblock In {\em Proc. IEEE Conference on Computer Vision and Pattern
  Recognition (CVPR)}, 2020.

\bibitem{li2018resound}
Yingwei Li, Yi Li, and Nuno Vasconcelos.
\newblock Resound: Towards action recognition without representation bias.
\newblock In {\em Proc. European Conference on Computer Vision (ECCV)}, 2018.

\bibitem{lin2019tsm}
Ji Lin, Chuang Gan, and Song Han.
\newblock Tsm: Temporal shift module for efficient video understanding.
\newblock In {\em Proc. IEEE International Conference on Computer Vision
  (ICCV)}, 2019.

\bibitem{liu2019learning}
Xingyu Liu, Joon-Young Lee, and Hailin Jin.
\newblock Learning video representations from correspondence proposals.
\newblock In {\em Proc. IEEE Conference on Computer Vision and Pattern
  Recognition (CVPR)}, 2019.

\bibitem{liu2020teinet}
Zhaoyang Liu, Donghao Luo, Yabiao Wang, Limin Wang, Ying Tai, Chengjie Wang,
  Jilin Li, Feiyue Huang, and Tong Lu.
\newblock Teinet: Towards an efficient architecture for video recognition.
\newblock In {\em Proc. AAAI Conference on Artificial Intelligence (AAAI)},
  2020.

\bibitem{loshchilov2016sgdr}
Ilya Loshchilov and Frank Hutter.
\newblock Sgdr: Stochastic gradient descent with warm restarts.
\newblock {\em arXiv preprint arXiv:1608.03983}, 2016.

\bibitem{luo2019grouped}
Chenxu Luo and Alan~L Yuille.
\newblock Grouped spatial-temporal aggregation for efficient action
  recognition.
\newblock In {\em Proc. IEEE International Conference on Computer Vision
  (ICCV)}, 2019.

\bibitem{materzynska2020something}
Joanna Materzynska, Tete Xiao, Roei Herzig, Huijuan Xu, Xiaolong Wang, and
  Trevor Darrell.
\newblock Something-else: Compositional action recognition with
  spatial-temporal interaction networks.
\newblock In {\em Proc. IEEE Conference on Computer Vision and Pattern
  Recognition (CVPR)}, 2020.

\bibitem{moon2021integralaction}
Gyeongsik Moon, Heeseung Kwon, Kyoung~Mu Lee, and Minsu Cho.
\newblock Integralaction: Pose-driven feature integration for robust human
  action recognition in videos.
\newblock In {\em Proc. IEEE Conference on Computer Vision and Pattern
  Recognition Workshops (CVPRW)}, 2021.

\bibitem{piergiovanni2019representation}
AJ Piergiovanni and Michael~S Ryoo.
\newblock Representation flow for action recognition.
\newblock In {\em Proc. IEEE Conference on Computer Vision and Pattern
  Recognition (CVPR)}, 2019.

\bibitem{ramachandran2019stand}
Prajit Ramachandran, Niki Parmar, Ashish Vaswani, Irwan Bello, Anselm Levskaya,
  and Jonathon Shlens.
\newblock Stand-alone self-attention in vision models.
\newblock {\em arXiv preprint arXiv:1906.05909}, 2019.

\bibitem{selvaraju2017grad}
Ramprasaath~R Selvaraju, Michael Cogswell, Abhishek Das, Ramakrishna Vedantam,
  Devi Parikh, and Dhruv Batra.
\newblock Grad-cam: Visual explanations from deep networks via gradient-based
  localization.
\newblock In {\em Proc. IEEE International Conference on Computer Vision
  (ICCV)}, 2017.

\bibitem{shao2020finegym}
Dian Shao, Yue Zhao, Bo Dai, and Dahua Lin.
\newblock Finegym: A hierarchical video dataset for fine-grained action
  understanding.
\newblock In {\em Proc. IEEE Conference on Computer Vision and Pattern
  Recognition (CVPR)}, 2020.

\bibitem{shechtman2007matching}
Eli Shechtman and Michal Irani.
\newblock Matching local self-similarities across images and videos.
\newblock In {\em Proc. IEEE Conference on Computer Vision and Pattern
  Recognition (CVPR)}, 2007.

\bibitem{simonyan2014two}
Karen Simonyan and Andrew Zisserman.
\newblock Two-stream convolutional networks for action recognition in videos.
\newblock In {\em Proc. Neural Information Processing Systems (NeurIPS)}, 2014.

\bibitem{srinivas2021bottleneck}
Aravind Srinivas, Tsung-Yi Lin, Niki Parmar, Jonathon Shlens, Pieter Abbeel,
  and Ashish Vaswani.
\newblock Bottleneck transformers for visual recognition.
\newblock {\em arXiv preprint arXiv:2101.11605}, 2021.

\bibitem{sudhakaran2020gate}
Swathikiran Sudhakaran, Sergio Escalera, and Oswald Lanz.
\newblock Gate-shift networks for video action recognition.
\newblock In {\em Proc. IEEE Conference on Computer Vision and Pattern
  Recognition (CVPR)}, 2020.

\bibitem{sun2018pwc}
Deqing Sun, Xiaodong Yang, Ming-Yu Liu, and Jan Kautz.
\newblock Pwc-net: Cnns for optical flow using pyramid, warping, and cost
  volume.
\newblock In {\em Proc. IEEE Conference on Computer Vision and Pattern
  Recognition (CVPR)}, 2018.

\bibitem{sun2018optical}
Shuyang Sun, Zhanghui Kuang, Lu Sheng, Wanli Ouyang, and Wei Zhang.
\newblock Optical flow guided feature: A fast and robust motion representation
  for video action recognition.
\newblock In {\em Proc. IEEE Conference on Computer Vision and Pattern
  Recognition (CVPR)}, 2018.

\bibitem{torabi2013local}
Atousa Torabi and Guillaume-Alexandre Bilodeau.
\newblock Local self-similarity-based registration of human rois in pairs of
  stereo thermal-visible videos.
\newblock {\em Pattern Recognition}, 46(2):578--589, 2013.

\bibitem{tran2015learning}
Du Tran, Lubomir Bourdev, Rob Fergus, Lorenzo Torresani, and Manohar Paluri.
\newblock Learning spatiotemporal features with 3d convolutional networks.
\newblock In {\em Proc. IEEE International Conference on Computer Vision
  (ICCV)}, 2015.

\bibitem{tran2019video}
Du Tran, Heng Wang, Lorenzo Torresani, and Matt Feiszli.
\newblock Video classification with channel-separated convolutional networks.
\newblock In {\em Proc. IEEE International Conference on Computer Vision
  (ICCV)}, 2019.

\bibitem{tran2018closer}
Du Tran, Heng Wang, Lorenzo Torresani, Jamie Ray, Yann LeCun, and Manohar
  Paluri.
\newblock A closer look at spatiotemporal convolutions for action recognition.
\newblock In {\em Proc. IEEE Conference on Computer Vision and Pattern
  Recognition (CVPR)}, 2018.

\bibitem{vaswani2017attention}
Ashish Vaswani, Noam Shazeer, Niki Parmar, Jakob Uszkoreit, Llion Jones,
  Aidan~N Gomez, Lukasz Kaiser, and Illia Polosukhin.
\newblock Attention is all you need.
\newblock {\em arXiv preprint arXiv:1706.03762}, 2017.

\bibitem{wang2020video}
Heng Wang, Du Tran, Lorenzo Torresani, and Matt Feiszli.
\newblock Video modeling with correlation networks.
\newblock In {\em Proc. IEEE Conference on Computer Vision and Pattern
  Recognition (CVPR)}, 2020.

\bibitem{wang2016temporal}
Limin Wang, Yuanjun Xiong, Zhe Wang, Yu Qiao, Dahua Lin, Xiaoou Tang, and Luc
  Van~Gool.
\newblock Temporal segment networks: Towards good practices for deep action
  recognition.
\newblock In {\em Proc. European Conference on Computer Vision (ECCV)}, 2016.

\bibitem{wang2018non}
Xiaolong Wang, Ross Girshick, Abhinav Gupta, and Kaiming He.
\newblock Non-local neural networks.
\newblock In {\em Proc. IEEE Conference on Computer Vision and Pattern
  Recognition (CVPR)}, 2018.

\bibitem{wang2018videos}
Xiaolong Wang and Abhinav Gupta.
\newblock Videos as space-time region graphs.
\newblock In {\em Proc. European Conference on Computer Vision (ECCV)}, pages
  399--417, 2018.

\bibitem{yang2020temporal}
Ceyuan Yang, Yinghao Xu, Jianping Shi, Bo Dai, and Bolei Zhou.
\newblock Temporal pyramid network for action recognition.
\newblock In {\em Proc. IEEE Conference on Computer Vision and Pattern
  Recognition (CVPR)}, 2020.

\bibitem{yang2019volumetric}
Gengshan Yang and Deva Ramanan.
\newblock Volumetric correspondence networks for optical flow.
\newblock In {\em Proc. Neural Information Processing Systems (NeurIPS)}, 2019.

\bibitem{zach2007duality}
Christopher Zach, Thomas Pock, and Horst Bischof.
\newblock A duality based approach for realtime tv-l 1 optical flow.
\newblock {\em Pattern Recognition}, pages 214--223, 2007.

\bibitem{zhao2020exploring}
Hengshuang Zhao, Jiaya Jia, and Vladlen Koltun.
\newblock Exploring self-attention for image recognition.
\newblock In {\em Proceedings of the IEEE/CVF Conference on Computer Vision and
  Pattern Recognition}, pages 10076--10085, 2020.

\bibitem{zhao2018trajectory}
Yue Zhao, Yuanjun Xiong, and Dahua Lin.
\newblock Trajectory convolution for action recognition.
\newblock In {\em Proc. Neural Information Processing Systems (NeurIPS)}, 2018.

\bibitem{zhou2018temporal}
Bolei Zhou, Alex Andonian, Aude Oliva, and Antonio Torralba.
\newblock Temporal relational reasoning in videos.
\newblock In {\em Proc. European Conference on Computer Vision (ECCV)}, 2018.

\bibitem{zolfaghari2018eco}
Mohammadreza Zolfaghari, Kamaljeet Singh, and Thomas Brox.
\newblock Eco: Efficient convolutional network for online video understanding.
\newblock In {\em Proc. European Conference on Computer Vision (ECCV)}, 2018.

\end{thebibliography}
